\let\latexdocument\document
\let\latexenddocument\enddocument

\documentclass[shortpaper]{clv3}
\let\document\latexdocument
\let\enddocument\latexenddocument

\makeatletter
\AtBeginDocument{%
  \if@filesw
    \immediate\openout\@mainqry=\jobname.qry
  \fi
}
\AtEndDocument{%
   \ifx\@biography\@empty\else{\par\ifbrief\vskip10pt\fi\biofont\noindent\@biography\par}\fi
   \immediate\closeout\@mainqry
      \ifodd\c@page\clearpage\thispagestyle{empty}\null\clearpage\else\clearpage\fi
}
\makeatother

\NewCommandCopy{\cnumdef}{\numdef}
\NewCommandCopy{\endcnumdef}{\endnumdef}
\let\numdef\relax \let\endnumdef\relax

\usepackage{linguex}

\usepackage{xcolor}
\usepackage{amsmath,amssymb,amsfonts}

\usepackage{adjustbox}
\usepackage{multirow}
\usepackage{makecell}
\usepackage{graphicx}
\usepackage{pgfplots}
\usepackage{subcaption}
\usepackage{float}
\usepackage{pgfplots}
\usepackage{stfloats}
\usepackage{hyperref}
\makeatletter
 \DeclareRobustCommand\ref{%
    \@ifstar\@refstar\T@ref
  }%
  \DeclareRobustCommand\pageref{%
    \@ifstar\@pagerefstar\T@pageref
  }%
 \makeatother
\newcommand{\mmodels}{\mathrel{||}\joinrel\Relbar}

\DeclareUnicodeCharacter{2212}{−}
\usepgfplotslibrary{groupplots,dateplot}
\usetikzlibrary{patterns,shapes.arrows}
\pgfplotsset{compat=newest}

\definecolor{darkblue}{rgb}{0, 0, 0.5}
\hypersetup{colorlinks=true,citecolor=darkblue, linkcolor=darkblue, urlcolor=darkblue}

\begin{document}


\runningtitle{Semantic Faithfulness of Transformer-based models}

\runningauthor{Chaturvedi et al.}


\title{Analyzing Semantic Faithfulness of Language Models via Input Intervention on Question Answering}


\author{Akshay Chaturvedi\thanks{E-mail: akshay91.isi@gmail.com.}}
\affil{IRIT, Universit\'{e} Paul Sabatier, Toulouse, France}

\author{Swarnadeep Bhar}
\affil{IRIT, Universit\'{e} Paul Sabatier, Toulouse, France}

\author{Soumadeep Saha}
\affil{Indian Statistical Institute, Kolkata, India}

\author{Utpal Garain}
\affil{Indian Statistical Institute, Kolkata, India}

\author{Nicholas Asher}
\affil{IRIT, Universit\'{e} Paul Sabatier, Toulouse, France}
\maketitle

\begin{abstract}
Transformer-based  language models have been shown to be highly effective for several NLP tasks. In this paper, we consider three transformer models, BERT, RoBERTa, and XLNet, in both small and large versions, and investigate how faithful their representations are with respect to the semantic content of texts. We formalize a notion of semantic faithfulness, in which the semantic content of a text should causally figure in a model's inferences in question answering. We then test this notion by observing a model's behavior on answering questions about a story after performing two novel semantic interventions\textemdash deletion intervention and negation intervention. While transformer models achieve high performance on standard question answering tasks, we show that they fail to be semantically faithful once we perform these interventions for a significant number of cases ($\sim 50\%$ for deletion intervention, and $\sim 20\%$ drop in accuracy for negation intervention). We then propose an  intervention-based training regime that can mitigate the undesirable effects for deletion intervention by a significant margin (from $\sim 50\%$ to $\sim 6\%$).  We analyze the inner-workings of the models to better understand the effectiveness of intervention-based training for deletion intervention. But we show that this training does not attenuate other aspects of semantic unfaithfulness such as the models' inability to deal with negation intervention or to capture the predicate\textendash argument structure of texts.  We also test InstructGPT, via prompting, for its ability to handle the two interventions and to capture predicate\textendash argument structure. While InstructGPT models do achieve very high performance on predicate\textendash argument structure task, they fail to respond adequately to our deletion and negation interventions.
\end{abstract}

\section{Introduction}

Transformer-based language models such as BERT~\cite{devlin-etal-2019-bert}, RoBERTa~\cite{roberta}, etc. have revolutionized natural language processing (NLP) research, generating contextualized representations that provide state of the art performance for various tasks like part of speech (POS) tagging, semantic role labelling etc.  The transfer learning ability of these models has discarded the need for designing task-specific NLP systems.  The latest incarnation of language models have now excited both the imagination and the fears of researchers \cite{black:etal:2022gpt,castelvecchi:2022} and journalists in the popular press; the models and chatbots based on them seem to be able to do code, argue and tell stories, but they also have trouble distinguishing fact from fiction.\footnote{Here is a sample of stories from the {\em New York Times}: `The New Chatbots Could Change the World. Can You Trust Them?' ({\em NYT} Dec. 10, 2022; `Meet GPT-3. It Has Learned to Code (and Blog and Argue)', {\em NYT}, Nov 24,2020; `The brilliance and the weirdness of ChatGPT', {\em NYT}, Dec. 5, 2022.}


\begin{table}
    \centering
    \begin{tabular}{l|l}
    \hline
        \textbf{Story} &  Once upon a time, in a barn    near a farm house, there lived \\ &  a little white kitten named  Cotton. Cotton lived high up  \\ &   [...] farmer's horses slept. \textbf{But Cotton wasn't alone in   } \textbf{her little }\\ &\textbf{home above the barn, oh no.}\\ \hline
          \textbf{Conversation }& What color was Cotton? white \\ \textbf{History }& Where did she live? in a barn \\ \hline
          \textbf{Question}& Did she live alone? \\ \hline
          \textbf{Prediction} & no \\ \hline
    \end{tabular}
    \caption{An example from CoQA dataset~\cite{10.1162/tacl_a_00266}. XLNet~\cite{xlnet} correctly predicts \emph{no} for the question \emph{``Did she live alone?"}. However, it still predicts \emph{no} when the rationale (i.e., text marked in bold) is removed from the story (i.e., deletion intervention).}
    \label{tab:del-inter}
\end{table}
Given their successes and their hold on the public imagination, researchers are increasingly interested in understanding the \emph{inner workings} of these models~\cite{liu-etal-2019-linguistic,tenney2018what,talmor-etal-2020-olmpics}.  In this paper, we look at how a fundamental property of linguistic meaning we call {\em semantic faithfulness} is encoded in the contextualized representations of transformer-based language models and how that information is used in inferences by the models when answering questions.  A semantically faithful model will accurately track the semantic content of questions and texts on which the answers to those questions are based. It is a crucial property for a model to have, if it is to distinguish facts about what is expressed in a text or conversation from fiction or hallucination.  We will show that current, popular transformer models are not semantically faithful.

This lack of semantic faithfulness highlights potential problems with popular language models trained with transformer architectures.  If these models are not semantically faithful, then they will fail to capture the actual semantic content of texts.  Operations that we develop in the body of the paper can be used to dramatically alter text content that these language models would not find, leading to errors with potentially important, negative socio\textendash economic consequences.   Even more worrisome is the instability that we have observed in these models and their occasional failure to keep predicate\textendash argument structure straight; if these models cannot reliably return information semantically entailed by textual content, then we can't rely on their predictions in many sensitive areas. Yet such systems are being deployed rapidly in these areas. 

In the next section, we discuss the virtues of semantic faithfulness and preview results of experiments that shed light on a model's semantic faithfulness.  In Section~\ref{sec:related}, we discuss the related work. In Section~\ref{sec:prelim}, we turn to the dataset and the transformer models that we will use in examining semantic faithfulness.  In Sections~\ref{sec:del} and \ref{sec:neg}, we introduce two types of interventions, deletion and negation interventions, that show that the language models lack semantic faithfulness.  In Section~\ref{sec:del}, we also discuss a kind of training that can help models acquire semantic faithfulness at least with respect to deletion intervention.   In Section~\ref{sec:pred}, we look at how models deal with predicate\textendash argument structure and with inferences involving semantically equivalent questions.  Once again we find that models lack semantic faithfulness. In Sections~\ref{sec:del},~\ref{sec:neg} and~\ref{sec:pred}, we also analyse semantic faithfulness of InstructGPT~\cite{gpt-3} via prompting. Finally, we conclude in Sections~\ref{sec:disc} and \ref{sec:conc}. The appendix section, i.e. Section~\ref{sec:appendix}, contains several examples of deletion and negation interventions.

\section{The fundamentals: semantic faithfulness} \label{sec:sem}
The property of interest is {\em semantic faithfulness}.  It relies on a basic theorem of all formal models of meaning in linguistics: the substitution of semantically equivalent expressions within a larger context should make no difference to the meaning of that context.
\DeclareRobustCommand{\mmodels}{\mathrel{||}\joinrel\Relbar}

\subsection{The definition of semantic faithfulness}

Linguists study meaning in the field of {\em semantics}, and {\em formal semantics} is the study of meaning using logical tools.  The basic tool of formal semanticists is the notion of a {\em structure} $\mathfrak{A}$ for a language ${\cal L}$ (${\cal L}$ structure) that assigns denotations of the appropriate type to all well-formed expressions of the language \cite{dowty:etal:1981}.  A denotation in an ${\cal L}$ structure for an ${\cal L}$ sentence $\phi$ is a truth condition, a formal specification of the circumstances in which $\phi$ is true. 

Formal semanticists use such structures to define {\em logical consequence} which allows them to study inferences that follow logically from the meaning of expressions.  $\phi$ of ${\cal L}$ is a {\em logical consequence} of a set of ${\cal L}$ sentences $S$ iff any ${\cal L}$ structure $\mathfrak{A}$ that makes all of $S$ true also makes $\phi$ true \cite{chang:keisler:1973}.  Formal semanticists equate logical consequence with {\em semantic entailment}, also known as {\em semantic consequence}, the notion according to which something follows from a text $T$ in virtue of $T$'s meaning alone.

Let $\mmodels$ denote the intuitive answerhood relation between a question $Q$ and answers $\phi, \psi$ to $Q$, where those answers follow from the semantic content of a story or text $T$ or model of its meaning $M_T$. Let $\models$ denote logical consequence or semantic entailment and let $\leftrightarrow$ denote the material biconditional of propositional logic.  A biconditional expresses an equivalence; so $\phi \leftrightarrow \psi$ means that $\phi$ will be true just in case $\psi$ is also true. For example, ``A goes to a party" $\leftrightarrow$ "B goes to a party" holds if and only if either both A and B or neither go to the party.\\  
{\bf Definition [Semantic faithfulness]}: If $T \models \phi \leftrightarrow \psi$, and $T \models Q \leftrightarrow Q'$, then $M_T$ is a semantically faithful model of $T$ iff:
\begin{equation}
 \label{sem}   
 T,Q \mmodels \phi \mbox{ iff } M_T,Q \mmodels  \psi
\end{equation}
and
 \begin{equation} \label{sem1}   
 M_T,Q \mmodels  \psi \mbox{ iff } M_T,Q' \mmodels  \psi
\end{equation}
Note that if $T \models Q \leftrightarrow Q'$ and $T \models \phi \leftrightarrow \psi$, then by the substitution of equal semantic values, it follows in formal semantics that $T,Q \mmodels \phi \mbox{ iff } T,Q' \mmodels  \psi$.  So in particular, this implies that $T,Q \mmodels \phi \mbox{ iff } M_T,Q \mmodels  \psi$. In the rest of the paper, we will concentrate on  the particular case of semantic faithfulness where $\phi = \psi$.

A semantically faithful machine learning model of meaning and question answering bases its answers to questions about $T$ on the intuitive, semantic content of $T$ and should mirror the inferences based on semantic consequence: if T's semantic content doesn't support an answer $\phi$ to question $Q$, then the model shouldn't provide $\phi$ in response to $Q$; if T's semantic content supports an answer $\phi$ to question $Q$, then the model should provide $\phi$ in response to $Q$.  Furthermore, if $T$ is altered to $T'$ so that 
while $T, Q \mmodels \psi$, $T', Q \not\mmodels \psi$, a semantically faithful model should replicate this pattern: $M_T, Q \mmodels \psi$, but $M_{T'}, Q \not\mmodels \psi$.  Thus, semantic faithfulness is a normative criterion that tells us how machine learning models of meaning should track human linguistic judgments, when textual input is altered in ways that are relevant to semantic meaning and semantic structure.

Linguistic meaning and the semantic consequence relation $\models$ are defined recursively over semantic structure.  Thus, semantic faithfulness provides an important window into machine learning models' grasp of semantic structure and its exploitation during inference.  In this semantic faithfulness goes far beyond and generalizes consistency tests based on lexical substitutions \cite{elazar-etal-2021-measuring}.  If a model is not semantically faithful, then it doesn't respect semantic entailment.  This in turn means that the model is not capturing correctly at least some aspects of semantic structure.  Semantic structure includes predicate\textendash argument structure (i.e. which object described in $T$ has which property) but also defines the scope of operators like negation over other components in the structure.  Semantic structure is an essential linguistic component for determining semantic faithfulness, since the semantic structure of a text is the crucial ingredient in recursively defining the logical consequence $\models$ relation that underlies semantic faithfulness.

To see how predicate\textendash argument structure links up with semantic faithfulness, suppose $T$ is the sentence ``{\em a blue car is in front of a red house".}  If $M_T$ does not respect predicate\textendash argument structure, then given the question ``{\em was the car red?"}, the model may reply yes, simply because it has seen the word ``red" in the text. Respecting predicate\textendash argument structure would dictate that the model predicts {\em no} for the question. Semantic structure also links semantic faithfulness with inference, as we exploit that structure to define valid inference.  The lack of semantic structure can cause the model to perform invalid inferences.  
 
 \subsection{A remark on language models and formal semantics}
 
Semantic faithfulness makes use of a traditional notion of logical consequence, which itself depends on truth conditional semantics, in which sentential meaning is defined in terms of the conditions under which the sentence is true.  So for instance, the meaning of {\em there is a red car} is defined in terms of the situations in which there is a car that is red \cite{davidson:1967}.  All this seems distant from the distributional view of semantics that informs language models (LMs).  However, the two are in fact complementary \cite{asher:2011}. \namecite{fernando:2004} shows how to provide a semantics of temporal expressions that fully captures their truth conditional meaning using strings of linguistic symbols or continuations.  \namecite{graf:2019} shows how to use strings to define generalized quantifiers,  which include the meanings of determiners like {\em every, some, finitely many} \cite{barwise:cooper:1981}. In such examples of continuation semantics, meanings are defined using strings of words or other symbols, similarly to distributional semantics, or functions over such strings.  When defined over the appropriate set of strings, continuation semantics subsumes can subsume the notion of logical consequence $\models$ under certain mild assumptions \citep{reynolds:1974}.  
 
 Inspired by this earlier literature,~\namecite{asher:etal:2017} provide a model of language in terms of a space of finite and infinite strings.  Many of these strings are just non meaningful sequences of words but the space also includes coherent and consistent strings that form meaningful texts and conversations.   Building on formal theories of textual and conversational meaning \cite{kamp:reyle:1993,asher:1993,degroote:2006,asher:pogodalla:2010}, \namecite{asher:etal:2017} use this subset of coherent and consistent texts and conversations to define the semantics and strategic consequences of conversational moves in terms of possible continuations. 

 Such continuation semantics also provides a more refined notion of meaning compared to that of truth conditional semantics.   Consider, for instance, the set of most probable continuations for \ref{hopeful}.  They are not the most probable continuations for \ref{disaster}, even though \ref{hopeful} and \ref{disaster} have the same truth conditional meaning.
 \ex.
 \a. If we do this, 2/3 of the population will be saved. \label{hopeful}
 \b. If we do this, 1/3 of the population will die. \label{disaster}

\noindent
\ref{hopeful}'s most likely continuations focus perhaps on implementation of the proposed action; \ref{disaster}'s most likely continuations would focus on a search for other alternatives.  Thus, while \ref{hopeful} and \ref{disaster} are semantically equivalent with respect to denotational, truth conditional semantics, they do not generate the same probability distribution over possible continuations and so have arguably distinct meanings in a continuation semantics that takes the future evolution of a conversation or text into account.  Thus, continuation semantics is a natural and non-conservative extension of truth conditional semantics. 

LMs find their place rather naturally in this framework \cite{fernando:2022}.  LMs provide a probability distribution over possible continuations.  Thus, they are sensitive to and can predict possible continuations of a given text or discourse.  LMs naturally can be seen as providing a continuation semantics.   In principle continuation semantics as practiced by LMs can in principle capture a finer grained semantics than denotational truth conditional semantics, as well as pragmatic and strategic elements of language.  

For an LM to be semantically faithful and to produce coherent texts and conversations, it must learn the right probability distribution over the right set of strings.  That is, it has to distinguish sense from nonsense, and it has then to recognize inconsistent from consistent strings, incoherent from coherent ones.  If it does so, then the LM will have mastered both $\models$ and the more difficult to capture notion of semantic coherence that underlies well-formed texts and conversations.  If it does not do so, it will not be semantically faithful.   The fact that continuation semantics subsumes the logical consequence relation $\models$ of formal and discourse semantics reinforces our contention that semantic faithfulness based on such a semantics should be a necessary constraint for adequate meanings based on continuations or meanings based on distributions. Semantic faithfulness is not only a test for an adequate notion of meaning but it also offers a road towards training LMs to better reflect an intuitive notion of semantic consequence and coherence.  

In designing experiments to test semantic faithfulness and LM model inference then, we need to pay attention to continuation semantics and how possible interventions can affect discourse continuations.  Our interventions exploit the semantics of continuations.  We need to do this, because LMs are sensitive to continuations and can detect low probablity continuations. Simple insertions of materials to affect semantic content threaten to not end up testing the inferences we want to test but rather signal an LM's sensitivity to low probability continuations.  Continuation semantics provides a rationale for human in the loop constructions of interventions that respect or shift semantic content and continuations \cite{kaushik:etal:2019,gardner:etal:2020evaluating}.

\subsection{A summary of our contributions}
We show that transformer representations of meanings are not semantically faithful, and this calls into question their grasp of semantic structure. We detail three types of experiments in which we show large language models fail to be semantically faithful: in the first case, transformers ``hallucinate'' responses to questions about texts that are not grounded in their semantic content; in the second, models fail to observe modifications of a text that renders it inconsistent with the model's answer to a question about the original text; in the third, we show that models don't reliably capture predicate\textendash argument structure. These are serious problems, as it means that we cannot offer guarantees that these sophisticated text understanding systems capture basic textual, semantic content. Hence, simple semantic inferences cannot be fully trusted.  We analyze the reasons for this and suggest some ways of remedying this defect.  

To investigate semantic faithfulness of a model $M$, we look at inferences $M$ must perform to answer a question, given a story or conversation $T$ and a conversation history containing other questions.  Table~\ref{tab:del-inter} shows an example. We look at question answering before and after performing two new operations, {\em deletion intervention} and {\em negation intervention}, that affect the semantic content of $T$.  We look at the particular case of the definition of semantic faithfulness where $\phi = \psi$---i.e. we look at whether the same ground-truth answer to a question is given before and after interventions.

\DeclareRobustCommand{\mmodels}{\mathrel{||}\joinrel\Relbar}
Deletion intervention removes from $T$ a text span conveying semantic information necessary and sufficient given $T$ for answering a question with answer $\psi$. We call the text conveying the targeted semantic information the {\em rationale.}  $T$ itself supports $\psi$ as an answer to $Q$---in the formalism of equation~\ref{sem}, $T, Q \mmodels \psi$.  But post intervention $T$, call it $d(T)$, does not: $d(T), Q \not\mmodels \psi$. The semantic content of $d(T)$ no longer semantically supports $\psi$.  A semantically faithful model $M_T$ should mirror this shift: $M_T, Q \mmodels \psi$ but $M_{d(T)} \not\mmodels \psi$---which accords with human practice and intuition.

\DeclareRobustCommand{\mmodels}{\mathrel{||}\joinrel\Relbar}
Negation intervention modifies a text $T$ into a text $n(T)$ such that $n(T)$ is inconsistent with $\psi$, where $\psi$ was an answer to a question supported by the original text.  In the formal terms we have used to defined semantic faithfulness, $T, Q \mmodels \psi$ but $n(T), Q \mmodels \neg\psi$.  One simple instance of negation intervention would insert a negation with scope over the $Q$ targeted semantic information.  But this is not the only or even the primary way; in fact such simple cases of negation intervention amount to only 10\% of our interventions. To preserve $T$'s discourse coherence and style, changing the content of a text so as to flip the answer in a yes/no questions typically requires other changes to $T$.  To consider a simple example, suppose that  in Table~\ref{tab:del-inter}, we consider as our question Q: {\em was Cotton white?} Performing negation intervention on the rationale, {\em  there lived
a little white kitten named Cotton}, led us to replace the rationale with two sentences: {\em there lived a little kitten named Cotton. Cotton was not white.}  

In general, negation intervention tests whether an ML model is sensitive to semantic consistency.  A semantically faithful model should no longer answer $Q$ with {\em yes} post negation intervention. Once again negation intervention exploits the notion of semantic faithfulness.  We should observe a shift in the ML model's behavior after negation intervention on a text $T$, $n(T)$: supposing that  on $T, Q \mmodels \psi$, a semantically faithful model $M_T$ should be such that $M_T, Q \mmodels \psi$ but $M_{n(T)} \not\mmodels \psi$.   

Deletion and negation interventions allow us to study the models' behavior in a counterfactual scenario.  Such counterfactual scenarios are  crucial to understanding the causal efficacy of the rationale in the models' inferring of the ground truth answer for a given question  \cite{scholkopf2019causality,kusner2017counterfactual,asher:etal:2021}.  Scientific experiments establish or refute  causal links between A and B by seeing what happens when A holds and what happens when $\neg A$ holds.  Generally, $A$ causes $B$ only if both A and B hold and the counterfactual claim, that if $\neg A$ were true then $\neg B$ would also be true, also holds.  So if we can show for intervention $i$ on $T$ that a model $M_T$ is such that $M_T, Q \mmodels \psi$ and $T, Q \mmodels \psi$ but also such that $M_{i(T)}, Q \mmodels \psi$ and $i(T), Q \not\mmodels \psi$---i.e., $i(T)$ no longer contains information $\alpha$ (originally in $T$) that is needed to support $\psi$ as an answer to $Q$---then we have shown that information $\alpha$ is not causally involved in the inference to $\psi$.




We perform our experiments on two question answering (QA) datasets, namely, CoQA~\cite{10.1162/tacl_a_00266}, and HotpotQA~\cite{yang2018hotpotqa}. CoQA is a conversational QA dataset whereas HotpotQA is a single-turn QA dataset.  Both the datasets include, for each question, an annotated rationale that human annotators determined to provide the ground truth answers to questions and from which ideally the answer should be computed in a text understanding system.  The {\bf bold text} in Table~\ref{tab:del-inter} is an example from CoQA of a rationale. We exploited these annotated rationales to study the language models' behavior under our semantic interventions. More precisely, we ask the following question: \emph{Do language models predict the ground truth answer even when the rationale is removed from the story under deletion intervention or negated under negation intervention?} 


The surprising answer to our question is ``yes". This shows that the models are not semantically faithful.  Intuitively, a model should not make such a prediction post deletion or negation intervention, since the content on which the ground truth answer should be computed, i.e. the rationale, is no longer present in the story. Our interventions show that the rationale is not a cause of the model's computing the ground truth answer; at least they are not necessary for computing the answer. This strongly suggests that such language models are not guaranteed to be semantically faithful, something we establish in greater detail in Sections \ref{sec:del} and \ref{sec:neg}.

In a third set of experiments in Section \ref{sec:pred}, we query the models directly for their knowledge of predicate\textendash argument structure in texts.  We construct sentences with two objects that each have a different property.  We then perform two experiments.  In the first, simple experiment, we simply query the model about the properties those objects have.  In some cases, some models had trouble even with this simple task.  
In a second set of experiments, we query the model with two distinct but semantically equivalent yes/ no questions. This experiment  produces some surprising results where models have trouble answering semantically equivalent questions in the same way, once again indicating a lack of semantic faithfulness.  Formally we have:
\begin{itemize}
    \item two questions, $Q, Q'$, 
    \item $T \models Q \leftrightarrow Q'$ and
    \item $T, Q \mmodels \psi \ \mathit{iff} \ T, Q' \mmodels \psi$
    \item but it's {\bf not} the case that\\ $M_T, Q \mmodels \psi  \ \mathit{iff} \ M_T, Q' \mmodels \psi$.
    \end{itemize}


Working with \emph{base} and \emph{large} variants of three language models, BERT~\cite{devlin-etal-2019-bert}, RoBERTa~\cite{roberta}, and XLNet~\cite{xlnet}, on the CoQA dataset~\cite{10.1162/tacl_a_00266}, we make the following five contributions:

\begin{enumerate}
    \item We show that, despite the models' high performance on CoQA and HotpotQA, they wrongly predict the ground truth answer post deletion intervention for a large number of cases ($\sim 50\%$). 
    \item We show that a simple intervention-based training strategy is extremely effective in making these models sensitive to deletion intervention without sacrificing high performance on the original dataset.  
    \item We quantitatively analyze the \emph{inner-workings} of these models by comparing the embeddings of common words under the two training strategies. We find that under intervention based training, the embeddings are more contextualized with regards to the rationale.
    \item For negation intervention, we show that all the models suffer a $\sim 20\%$ drop in accuracy when the rationale is negated in the story. 
    \item  We show that, in general,  the models have difficulty in capturing predicate\textendash argument structure by examining their behavior on paraphrased questions. 
    
    \item We also test the ability of InstructGPT~\cite{gpt-3} (i.e. \emph{text-davinci-002} and \emph{text-davinci-003}) to tackle the two interventions and capture predicate\textendash argument structure via prompting. For the two interventions, InstructGPT models also displays similar behavior as the other models. With regards to predicate\textendash argument structure, the models achieves very high performance. However, for certain cases, the models do exhibit inconsistent behavior as detailed in Section~\ref{sec:pred}.
\end{enumerate}

\section{Related work} \label{sec:related}

There has been a significant amount of research analyzing language models' behavior across different NLP tasks~\cite{rogers-etal-2020-primer}.  \emph{Probing} has been a popular technique to investigate linguistic structures encoded in the contextualized representations of these models~\cite{pimentel-etal-2020-information,hewitt-liang-2019-designing,hewitt-manning-2019-structural,chi:etal:2020}.  In probing, one trains a model (known as a \emph{probe}) which takes the frozen representations of the language model as input, for a particular linguistic task. The high performance of the probe implies that the contextualized representations have encoded the required linguistic information.

In particular, predicate\textendash argument structure has been a subject of probing~\cite{conia2020bridging,conia2022probing}. Though most, if not all, of the effort is devoted to finding arguments of verbal predicates denoting actions or events using semantic role labeling formalisms~\cite{chi:etal:2020,conia2020bridging}.  Little effort has been made in the literature to investigate the grasp of predicate\textendash argument structure at the level of the formal semantic translations of natural language text---which includes the arguments of verbal predicates but also things like adjectival modification.

One major disadvantage of probing methods is that they fail to address how this information is used during inference~\cite{tenney-etal-2019-bert,rogers-etal-2020-primer}. Probing only shows that there are enough clues in the representation so that a probe model can learn to find, say the predicate\textendash argument from the language model's representation.  It tells us little as to whether the model leverages that implicit information in reasoning about textual content.  Our experiments are designed to do the latter.


Another approach to understanding the inner workings of language models studies their behavior at inference time. ~\namecite{amnesic} explores an intervention-based model analysis, called \emph{amnesic probing}. Amnesic probing performs interventions on the hidden representations of the model in order to remove specific morphological information.  In principle one could extend this approach to other kinds of linguistic information.  Amnesic probing is unlike our work, in which the interventions are performed on the input linguistic content and form.  \namecite {balasubramanian-etal-2020-whats} showed in related work that BERT is \emph{surprisingly brittle} when one named entity is replaced by another.~\namecite{DBLP:journals/corr/abs-2003-04985} showed the lack of robustness of BERT to commonly occurring misspellings.  

For the task of question answering, a Transformer-based language model with multiple output heads is typically used~\cite{hu-etal-2019-multi}. An output head caters to a particular \emph{answer type}. Thus, the usage of multiple output heads allows the model to generate different answer types such as span, yes/no, number, etc.~\namecite{geva-etal-2021-whats} studied the behavior of \emph{non-target} heads, i.e., output heads not being used for prediction. They showed that, in some cases, non-target heads are able to explain the models' prediction generated by the \emph{target head}.~\namecite{schuff-etal-2020-f1} analyzed the question answering models which predict answer as well as an explanation. For such models, they manually analyzed the predicted answer and explanation to show that the explanation is often not suitable for the predicted answer. Their methodology is in contrast to our work, since we simply argue that the model uses the rationale for predicting the answer if it is sensitive to \emph{deletion intervention}.

Researchers in prior work have also studied the behavior of the model on manipulated input texts~\cite{balasubramanian-etal-2020-whats,DBLP:journals/corr/abs-2003-04985,jia-liang-2017-adversarial,song-etal-2021-universal,belinkov2018synthetic,zhang:etal:2020}. However, they usually frame the task in an \emph{adversarial scenario} and rely either on an attack algorithm or complex heuristics for generating manipulated text. The objective in such work is to fool the model with the manipulated text so that the model changes its predictions whereas a human would not change the prediction in the face of the manipulated data.  

In contrast, deletion intervention is a simple \emph{content deletion} strategy; it is not designed to get the model to shift its predictions in cases where a human would not.  It's not designed to trick or fool ML models.  Deletion intervention manipulates the text to test how the deletion of content affects inference; ideally both humans and the ML model should shift their predictions in a similar way given a deletion intervention. Nevertheless, it is also reasonable to expect a model that was successfully attacked in an adversarial setting to be sensitive to deletion intervention. 

With respect to negation intervention, researchers have examined the effects of negation and inference at the sentential level on synthetic datasets  \cite{naik-etal-2018-stress, kassner-schutze-2020-negated, hossain-etal-2020-analysis, hosseini-etal-2021-understanding}.  Our aim is more ambitious; we study how transformer models encode both the content $C$ in a text and content $C'$ in a negation-intervened text that is inconsistent with $C$.  Using negation intervention, we test how replacing $C$ with $C'$ affects inference in natural settings.  As with deletion intervention, we offer another way of changing the meaning of texts that should make both humans and semantically faithful models change their predictions.  There is similar work relevant to negation intervention---on contrast set data and also counterfactual data~\cite{kaushik:etal:2019,gardner:etal:2020evaluating}.  The datasets on which~\namecite{kaushik:etal:2019,gardner:etal:2020evaluating} operate are less complex discursively than the CoQA dataset.  Generally, our tests also go beyond tests of consistency as in~\namecite{elazar-etal-2021-measuring}, which is based on lexical substitutions.  Semantic faithfulness, which we test with deletion and negation interventions as well as questions paraphrases, works at a structural level in semantics; it generalizes~\namecite{elazar-etal-2021-measuring}'s notion of consistency. Also, while~\namecite{elazar-etal-2021-measuring} focused on pretrained language models, our work is focused on language models finetuned for question answering.

Our study on CoQA and HotpotQA also allows us to look more closely at what the models are actually sensitive to in a longer text or story.  We return to this issue in more detail in Section~\ref{sec:neg}.   In general, interventions are an important mechanism to build counterfactual models as~\namecite{kaushik:etal:2019} also argue.  These are important for understanding causal structure \cite{scholkopf2019causality, kusner2017counterfactual, barocas-hardt-narayanan}.

\section{Dataset Specification and Model Architecture}\label{sec:prelim}
We now describe the two datasets: CoQA and HotpotQA; and the architecture of the models used for this work along with implementation details.
\subsection{Dataset}

\begin{table}[t]
\centering
\small
\begin{tabular}{|c|c|c|c|c|}
\hline
Dataset & Split & Story & Questions & unk\%\\
\hline
\multirow{2}{*}{CoQA}    & train & 7199 & 108647 & 1.26\\
                         & dev & 500 & 7983   & 0.83 \\
\hline
\multirow{2}{*}{HotpotQA}   & train & 84579  & 90447 & - \\
                            & dev  & 7350 & 7405  & - \\ 
\hline
\end{tabular}
\caption{Data Statistics for CoQA and HotpotQA along with percentage of \emph{unknown} questions}
\label{tab:dataset}
\end{table}
The CoQA dataset consists of a set of stories paired with a sequence of questions based on the story. To answer a particular question, the model has access to the story and previous questions with their ground truth answers---this is the conversation history. The dataset contains questions of five types: \emph{yes/no} questions, questions whose direct answer is a \emph{number}, alternative or \emph{option} questions (e.g., {\em do you want tea or coffee?}), questions with an \emph{unknown} answer, and questions whose answer is contained in a \emph{span} of text. The \emph{span} answer type accounts for majority of the questions ($>75\%$). The dataset also contains a \emph{human annotated rationale} for each question. 


The HotpotQA is a standard (i.e. single-turn) QA dataset. For each question, the dataset contains $2$ gold, and $8$ distractor wikipedia paragraphs. Only the gold paragraphs are relevant to the question. The annotated rationale highlights the particular sentences within the gold paragraphs which are needed to answer the question. Given that the language models used in this work have an input limit of $512$ tokens, we only feed the two gold paragraphs as input to the model. For the sake of consistency with CoQA, we refer to this concatenated input as \emph{story}. HotpotQA mostly contains of span answer type questions ($>90\%$) and unlike CoQA, this dataset doesn't contain any question with an \emph{unknown} answer.  Since the test set for the two datasets is not publicly available, we report the performance of the models across different experimental settings on the development set. Table~\ref{tab:dataset} provides statistics for training and development set for the two datasets.

\subsection{Models}\label{sec:method}

We conducted experiments on \emph{base} and \emph{large} variants of three Transformer-based language models\textemdash BERT~\cite{devlin-etal-2019-bert}, RoBERTa~\cite{roberta} and XLNet~\cite{xlnet}. For the sake of consistency, we used the same model architecture for the two datasets. The only difference lies in the input given to the model. For CoQA, to predict the answer for the $i^{th}$ question, $Q_{i}$, for a given story $S$, the models use previous questions along and their ground truth answers from the conversation history. Therefore, the input for the models for the story, $S$, and question, $Q_{i}$, is as follows.  
\begin{align*}
    \textrm{XLNet : } & [S \textrm{ <sep> } Q_{i-2} A_{i-2}  Q_{i-1} A_{i-1} \\& Q_i \textrm{ <sep> } \textrm{ <cls> } ] \\
    \textrm{BERT/RoBERTa : }  & [ \textrm{ <cls> } Q_{i-2} A_{i-2} Q_{i-1} A_{i-1} Q_i\\     & \textrm{ <sep> } S \textrm{ <sep> }  ] \\
\end{align*}
\noindent
where <sep> token is used to demarcate the story and the question history, <cls> is a special token which is used for predicting answer for non-span type questions, and $A_{j}$ denotes the ground truth answer for the question, $Q_{j}$ . In the rest of the paper, we refer to the string $Q_{i-2} A_{i-2}  Q_{i-1} A_{i-1} Q_{i}$ as \emph{question context}. Since HotpotQA is a single-turn QA dataset, instead of feeding the question content, we only feed the current question $Q_{i}$ to the model. The rest of the input remains the same.

We used the publicly available XLNet model for this paper\footnote{\url{https://github.com/stevezheng23/mrc_tf}}. The model contains output heads for \emph{unknown}, \emph{yes}, \emph{no}, \emph{number}, \emph{option}, and \emph{span}. Each output head is fed with a concatenation of the CLS embedding and contextualized embeddings of the story weighted by the predicted start probabilities to predict a score.

For BERT and RoBERTa, we implemented the rationale tagging multi-task model described in~\namecite{roberta:19}. Unlike XLNet, the two models are trained on question answering as well as on the rationale tagging task. Furthermore, for a question, the two models can predict \emph{yes}, \emph{no}, \emph{unknown}, and \emph{span}. As a result, for CoQA, the two models predict span for $78.9\%$  of the questions in the development set, whereas XLNet predicts span for $75.8\%$. Similarly, for HotpotQA, the two models predict span for $93.8\%$ of questions in the development set, whereas XLNet predicts span for $91.1\%$. For span prediction, the start and end logits for the answer are predicted by applying a fully connected layer to the contextualized representation of the story obtained from the last layer of the model.

The rationale tagging task requires predicting whether a token $t \in S$ belongs to the rationale. Let $h_{t} \in \mathbb{R}^{d}$ denote the contextualized embedding obtained from the last layer for token $t$. The model assigns a probability $p_{t}$ for $t$ to be in the rationale as follows.

\begin{equation}\label{eq:1}
    p_{t} = \sigma(u ReLU(V h_{t}))
\end{equation}
\noindent
where $u\in \mathbb{R}^{1\times d}$, $V\in \mathbb{R}^{d\times d}$, $ReLU$ is the rectified linear unit activation function, and $\sigma(.)$ denotes the sigmoid function. An attention mechanism is then used to generate a representation, $q^{L}$, as shown below.

\begin{align}\label{eq:2}
    p'_{t} =  p_{t} \times h_{t} & \\
    a_{t} = softma&x(w_{1}ReLU(W_{2} p'_{t})) \\
    q^{L} = \sum_{t} a_{t} \times p'_{t}&
\end{align}
where $w_{1}\in \mathbb{R}^{1\times d}$, $W_{2}\in \mathbb{R}^{d\times d}$. Let $h_{CLS} \in \mathbb{R}^{d}$ denote the CLS embedding obtained from the last layer. $h_{CLS}$ is concatenated with the embedding $q^{L}$. The concatenated embedding is then used in BERT and RoBERTa to generate a score for \emph{yes}, \emph{no}, and \emph{unknown} respectively.

\subsection{Implementation details}\label{sec:impl-details}
We implemented the three language models in PyTorch using the Huggingface library~\cite{wolf-etal-2020-transformers}. The models were finetuned on the CoQA dataset for $1$ epoch. The \emph{base} variant of the three models  was trained on a single $11$ GB GTX $1080$ Ti GPU, whereas the \emph{large} variant was trained on a single $24$ GB Quadro RTX $6000$ GPU. The code and the additional datasets created as part of this work are publicly available\footnote{\url{https://github.com/akshay107/sem-faithfulness}}.

\begin{table}[t]
    \small
    \centering
    \begin{tabular}{l|l}
    \hline
        \textbf{Story} &  Characters: Sandy, Rose, Jane, Justin, Mrs. Lin \\ &
  [...]  \\ &  Jane: Sandy, I called you yesterday. Your mother told me \\ & [...] This year is very important to us.  \\ & \textbf{Sandy:(Crying) My father has lost his job}, and we have  \\ &  no money to pay all the spending.   \\ & [...]  \\ &  \textit{Jane: Eh...I hear that Sandy's father has lost his job}, and  \\ & Sandy has a part-time job. \\\hline
          \textbf{Question}& Who was unemployed? \\ \hline
          \textbf{Prediction} & Sandy's father \\ \hline
    \end{tabular}
    \caption{An example from CoQA dataset where the rationale (shown in bold) is not necessary to answer the question. The question can be answered using the italicised text.}
    \label{tab:truncstory}
\end{table}

\section{Deletion Intervention and Results}\label{sec:del}

In this section, we explain the operation of \emph{deletion intervention} and discuss the proposed intervention-based training. Deletion intervention is an operation that removes the \emph{rationale} of a question $Q$ from the \emph{story}. For a few instances in the CoQA dataset, we found that the annotated \emph{rationale} for $Q$ was not necessary for answering Q, because the sentences following the rationale contained the relevant information for supplying an answer to $Q$. One such instance is shown in Table~\ref{tab:truncstory}.  In our experiments with CoQA, we did not find any instance where the sentences preceding the rationale contained the necessary information for answering the question. To avoid problems with such examples containing redundancies, given an original story (OS), we created three additional datasets:

\begin{enumerate}
    \item TS: In this dataset, we truncate the original story (OS) so that the statement containing the rationale is the last statement. We refer to this dataset as TS (short for \emph{truncated story}). The stories in TS do not reduplicate elsewhere information in the rationale. TS has same number of samples as OS.
    \item TS-R: Given TS, we perform \emph{deletion intervention} by removing all the sentences containing the rationale. The cases where the rationale begins from the first sentence itself are discarded. For questions where the model predicts a  \emph{span}, we add the ground truth answer (if not already present) post deletion intervention. This is necessary since for the \emph{span} type questions, the model can only predict the ground truth answer if it is present in the story. As an example, consider the question ``Where does Alan go after work?" and the story ``Alan works in an office. \textbf{He goes to a nearby park after work.}" (rationale shown in bold). In this case, TS-R will be ``Alan works in an office. park." Since TS-R doesn't contain the information necessary for answering the question, the model should predict \emph{unknown} for such instances. TS-R had total of $98436$ samples for training and $7271$ samples for evaluation.
    \item TS-R+Aug: This dataset is similar to TS-R. However, in this dataset, instead of simply adding the ground truth answer at the end for the aforementioned cases, we use the OpenAI API (i.e., \emph{gpt-3.5-turbo}) to generate a sentence containing the ground truth answer. This generated sentence is then added at the end for the concerned cases. Given the budgetary constraints involved in using the OpenAI API, we only use this dataset for evaluation, and not during training. Generating sentence using the API is not essential for deletion intervention. Rather, the purpose of this dataset is to show that the proposed intervention-based training doesn't solely rely on superficial cues of TS-R to tackle deletion intervention. Out of the $5351$ cases of span questions where ground truth answer was absent post deletion intervention, the API was successfully able to generate sentence containing the ground truth answer for $4329$ cases. Unsuccessful cases were discarded from this dataset. Similar to TS-R, the model should predict \emph{unknown} for TS-R+Aug. This dataset had total of $6249$ samples for evaluation.
\end{enumerate}

\begin{table}[t]
\centering
\small
\begin{tabular}{|c|c|c|c|c||c|c|c|}
\hline
\multirow{2}{*}{Model} & \multirow{2}{*}{Dataset} & \multicolumn{3}{c||}{OT} & \multicolumn{3}{c|}{IBT}\\
\cline{3-8}
& & F1 & EM & unk\%& F1 & EM & unk\%\\
\hline
\multirow{3}{*}{BERT-base}    & OS & 76.1 & 66.3 & 1.97 & 76.4 & 67.2 & 3.82\\
        &TS & 77.2 & 67.1 & 2.18 & 77.7 & 68.0 & 7.93\\
        &TS-R & 55.6 & 48.2 & 1.98 & 5.7 & 5.4 & 93.08\\
        &TS-R+Aug & 51.5 & 44.3 & 7.10 & 42.4 & 36.3 & 45.50\\
\hline
\multirow{3}{*}{BERT-large} &OS & 80.7 &  71.1 & 2.01 & 78.8 &  69.8 & 4.20\\
        &TS &    81.6 &  72.1 & 2.32 & 80.1 &  70.7 & 7.34\\
        &TS-R &    63.6 &  57.8 & 3.79 & 5.4 &  5.1 & 94.25\\
        &TS-R+Aug & 53.4 & 46.3 & 8.80  & 38.3 & 32.9 & 51.88\\
\hline
\multirow{3}{*}{RoBERTa-base} &OS & 80.3 &  70.8 & 1.95 & 81.2 &  71.6 & 2.86\\
        &TS &    80.8 &  71.1 & 2.64 & 81.9 &  72.0 & 5.20\\
        &TS-R &    55.5 &  51.1 & 16.92 & 5.5 &  5.3 & 94.25\\
        &TS-R+Aug & 39.2 & 28.2 & 14.26 & 6.3 & 6.0 & 92.13\\
\hline
\multirow{3}{*}{RoBERTa-large} &OS & 87.0 &  77.7 & 1.74 & 86.2 &  76.9 & 2.66\\
        &TS &    86.8 &  77.3 & 2.72 & 86.3 &  76.7 & 4.01\\
        &TS-R &    59.9 &  55.7 & 22.36 & 5.1 &  5.0 & 95.34\\
        &TS-R+Aug & 42.9 & 32.0 & 22.18 & 6.0 & 5.7 & 93.65\\
\hline
\multirow{3}{*}{XLNet-base} & OS & 82.5 &  74.8 & 1.08 & 81.3 &  74.2 & 4.63\\
      & TS &    82.1 &  74.2 & 1.11 & 79.6 &  72.4 & 10.87\\
      & TS-R &    53.5 &  48.0 & 14.00 & 6.6 &  6.4 & 93.86\\
      &TS-R+Aug & 50.7 & 45.3 & 13.45 & 25.2 & 22.2 & 68.75\\
\hline
\multirow{3}{*}{XLNet-large} & OS & 86.3 &  78.9 & 0.86 & 83.1 &  75.8 & 5.10\\
      & TS &    85.6 &  78.5 & 2.58 & 81.0 &  74.1 & 10.69\\
      & TS-R &    48.1 &  44.3 & 31.68 & 5.6 &  5.5 & 95.42\\
      &TS-R+Aug & 46.9 & 42.4 & 26.18 & 22.1 & 19.5 & 73.93\\
\hline
\end{tabular}
\caption{EM, F1 score, and percentage of unknown predictions (i.e., $unk\%$) of the models under the two training strategies for CoQA.}
\label{tab:OT-IBT}
\end{table}

\begin{table}
\centering
\small
\begin{tabular}{|c|c|c|c|c||c|c|c|}
\hline
\multirow{2}{*}{Model} & \multirow{2}{*}{Dataset} & \multicolumn{3}{c||}{OT} & \multicolumn{3}{c|}{IBT}\\
\cline{3-8}
& & F1 & EM & unk\%& F1 & EM & unk\%\\
\hline
\multirow{3}{*}{BERT-base}    & OS & 72.3 & 56.7 & 0.16 & 71.2 & 55.8 & 1.00 \\
        &OS-R & 59.5 & 48.5 & 4.60 & 0.4 & 0.3 & 99.05\\
        &OS-R+Aug & 63.1 & 52.2 & 4.27 & 3.1 & 2.2 & 93.96\\
\hline
\multirow{3}{*}{BERT-large} & OS &  74.8 &  59.6 & 0.21 & 73.8 & 58.5 & 1.24\\
        &OS-R & 63.7 &  53.8 & 5.63 & 0.5 & 0.4 & 99.17\\
        &OS-R+Aug & 64.7 & 54.2 & 5.36 & 2.63 & 2.05 & 95.46\\
\hline
\multirow{3}{*}{RoBERTa-base} &OS & 72.0 &  56.7 & 0.16 & 72.7 & 57.4 & 0.73\\
        &OS-R &  66.2 &  59.1 & 0.86 & 0.6 & 0.5 & 98.86\\
        &OS-R+Aug & 36.9 & 15.7 & 0.94 & 0.9 & 0.6 & 97.93\\
\hline
\multirow{3}{*}{RoBERTa-large} &OS & 80.0 &  64.5 & 0.18 & 79.7 & 64.4 & 0.70\\
        &OS-R &   75.2 &  70.0 & 2.84 & 0.6 & 0.5 & 99.06\\
        &OS-R+Aug & 40.3 & 18.4 & 3.90 & 0.9 & 0.6 & 97.86\\
\hline
\multirow{3}{*}{XLNet-base} & OS & 74.2 &  60.1 & 0.07 & 73.5 & 59.4 & 1.05\\
      & OS-R &   63.0 &  53.0 & 0.73 & 0.6 & 0.4 & 98.85\\
      & OS-R+Aug & 63.5 & 53.9 &  1.16 & 3.1 & 2.4 & 94.30\\
\hline
\multirow{3}{*}{XLNet-large} & OS & 80.0 &  66.1 & 0.23 & 77.4 & 63.5 & 1.03\\
      & OS-R &  68.5 &  59.1 & 0.60  & 0.4 & 0.3 & 99.21\\
      & OS-R+Aug & 62.7 & 53.7 & 9.12 & 1.6 & 1.1 & 96.81\\
\hline
\end{tabular}
\caption{EM, F1 score, and percentage of unknown predictions (i.e., $unk\%$) of the models under the two training strategies for HotpotQA.}
\label{tab:Hotpot-OT-IBT}
\end{table}

\begin{table}[t]
\centering
\footnotesize
\begin{tabular}{|c|c|c|c|c|}
\hline
Model & \multicolumn{2}{c|}{Dataset} & EM & F1 \\
\hline
\multirow{4}{*}{text-davinci-002} & \multirow{2}{*}{CoQA} &TS-R  & 46.4 & 58.5   \\
 & & TS-R+Aug  & 40.4 & 53.3   \\ \cline{2-5}
 & \multirow{2}{*}{HotpotQA} & OS-R  & 14.1 & 32.2   \\
& & OS-R+Aug  & 11.4 & 29.5   \\
\hline
\multirow{4}{*}{text-davinci-003} & \multirow{2}{*}{CoQA} & TS-R   & 28.0 & 45.6  \\
& & TS-R+Aug & 17.3 & 34.9  \\ \cline{2-5}
& \multirow{2}{*}{HotpotQA} & OS-R   & 25.6 & 41.7  \\
& & OS-R+Aug & 18.0 & 34.1  \\
\hline
\end{tabular}
\caption{Deletion Intervention: EM and F1 score for the two InstructGPT models.}
\label{tab:gpt3-delintnew}
\end{table}
For HotpotQA, given an original story (OS), we constructed two additional datasets: OS-R, OS-R+Aug in a similar fashion. Unlike Table~\ref{tab:truncstory}, we did not find any such problematic cases for HotpotQA. For OS-R+Aug, out of $5855$ cases of span type questions where the the ground truth answer was absent post deletion intervention, \emph{gpt-3.5-turbo} was successful in generating a sentence containing the ground-truth answer for $5457$ cases. Similar to TS-R+Aug, OS-R+Aug dataset is also used solely for evaluation. OS-R had total of $87234$ samples for training and $7119$ samples for evaluation. OS-R+Aug had total of $6721$ samples for evaluation.

We trained the models on the OS dataset and evaluated them on the aforementioned datasets. We refer to this training strategy as OT (short for original training). The OT in Tables~\ref{tab:OT-IBT} and~\ref{tab:Hotpot-OT-IBT} shows EM (exact match), F1, and the percentage of unknown predictions (unk $\%$) of the models for this training strategy on CoQA and HotpotQA respectively. Note that, for all the datasets, the EM, and F1 in the two tables is with respect to the original ground-truth answer (i.e., ground truth answer for OS). As we can see from OT in Table~\ref{tab:OT-IBT}, for the datasets OS and TS, the performance of all the models is pretty similar. The performance drops for TS-R which shows some sensitivity to \emph{deletion intervention}. However, all the models still achieve an EM of $\sim 50\%$ which is intuitively way too high for a semantically faithful model.  We believe this shows that the models rely on superficial cues for predicting the answer;  for example, in the presence of a question like ``What color was X?" it searches for a color word not too far from a mention of X.  We also find that, for TS-R, the unk $\%$ for RoBERTa, XLNet is significantly higher than BERT. There is a drop in EM, and F1 for TS-R+Aug in comparison to TS-R (especially for RoBERTa) but the unk $\%$ remains similar which shows that the model is not adept at handling this dataset. For HotpotQA, the OT in Table~\ref{tab:Hotpot-OT-IBT} highlights the models' inability to tackle deletion intervention even further. We can see that for OS and OS-R, the performance drop is not so apparent. In fact, RoBERTa actually achieves a higher EM on OS-R than OS. Overall, all the models have very high scores on OS-R which is undesirable. For OS-R+Aug, we see that BERT, and XLNet-base have higher scores in comparison to OS-R. Whereas, the scores drop for the other models like RoBERTa. However, like CoQA, the  unk $\%$ remains similar.

Apart from the six models, we also evaluated the behavior of InstructGPT~\cite{gpt-3} using OpenAI API on deletion intervention. We looked at two InstructGPT models: \emph{text-davinci-002} and \emph{text-davinci-003}. As per the models' documentation~\footnote{\url{https://platform.openai.com/docs/models/gpt-3-5}}, the two models are pretty similar. The only difference being that \emph{text-davinci-002} was trained with supervised learning instead of reinforcement learning. For CoQA, we provided the story, the question context (i.e. two previous question with their ground truth answer), and the current question as input prompt. Whereas, for HotpotQA, only the story and the current question is given as input prompt. We calculated the EM and F1 score for the two models on TS-R, TS-R+Aug datasets for CoQA and OS-R, OS-R+Aug datasets for HotpotQA. Table~\ref{tab:gpt3-delintnew} shows these scores. For CoQA, we see that \emph{text-davinci-002} achieves very high scores; for nearly $\sim 50\%$ of cases the model continues to predict the ground truth answer for TS-R. This pathological behavior is similar to other models studied in this work. While \emph{text-davinci-003} does achieve lower scores for TS-R, it sill predicts the ground truth answer for $\sim 30\%$ of the cases. The scores for both the models are lower for TS-R+Aug than TS-R. Unlike CoQA, for HotpotQA, \emph{text-davinci-002} achieves lower scores than \emph{text-davinci-003} on OS-R. We see that \emph{text-davinci-003} predicts ground truth answer for $25.6\%$ of the cases of OS-R. The scores for both the models are lower on OS-R+Aug than OS-R. Overall, these result show that InstructGPT models do not respond appropriately to deletion intervention. We provide several examples of deletion intervention in Section~\ref{sec:appendix}.

\subsection{Intervention-based Training}\label{sec:ibt}


To enhance the sensitivity of the language models to deletion intervention, we propose a simple \emph{intervention-based training} (IBT). In this training strategy, we train the model on multiple datasets simultaneously. For CoQA, the model is trained to predict the ground truth answer for OS and TS, whereas, for TS-R, the model is trained to predict \emph{unknown}. Similarly, for HotpotQA, the model is trained to predict the ground truth answer for OS and to predict \emph{unknown} for OS-R. Note that the models are trained for same number of epochs under both the training strategies. The IBT in Tables~\ref{tab:OT-IBT} and~\ref{tab:Hotpot-OT-IBT} shows EM (exact match), F1, and the percentage of unknown predictions (unk $\%$) of the models for this training strategy on CoQA and HotpotQA respectively. Firstly, we observe that for the datasets OS and TS of CoQA and OS of HotpotQA, the training strategy IBT is at par with the strategy OT. Also, we see that the performance of the models drops significantly and all the models also have very high unk $\%$ ($> 90\%$) post deletion intervention (TS-R for CoQA and OS-R for HotpotQA). Thus, IBT is able to make the models highly sensitive to \emph{deletion intervention}. Furthermore, for OS-R+Aug of HotpotQA, the models' performance is similar to OS-R under IBT. For TS-R+Aug of CoQA, RoBERTa and XLNet have very high unk $\%$. While the unk $\%$ for BERT drops significantly, IBT still fares better than OT on TS-R+Aug dataset. This shows that the models trained under IBT do not solely rely on superficial cues from TS-R (or OS-R) to predict \emph{unknown}.



\subsection{In-depth Analysis of Intervention-based Training}

\begin{figure}
\centering
\scalebox{.5}{\input{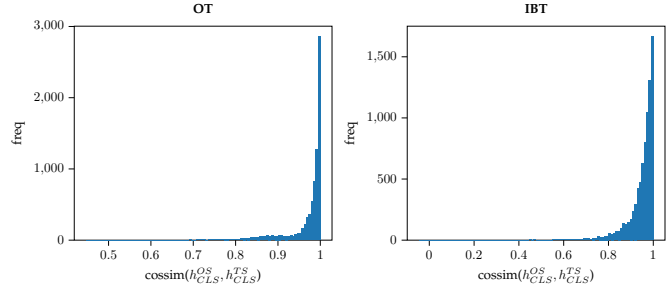}}
\caption{RoBERTa-large: Histogram plot of cosine similarity between CLS embedding for OS and TS under two training strategies (OT on left and IBT on right).}
\label{fig:cls-1}
\end{figure}

In this section,  we study the inner-workings of these models in order to explain the effectiveness of intervention-based training against deletion intervention. As mentioned in \S~\ref{sec:method}, CLS embedding plays a crucial role in predicting an answer to a particular question. Hence, to begin with, we look at the cosine similarity ({\em cossim}) between CLS embeddings of OS and TS under the two training strategies (OT and IBT). Similarly, we also look at the cossim between CLS embeddings of OS and TS-R under the two training strategies. Figures~\ref{fig:cls-1} and~\ref{fig:cls-2}  show the histogram of cossim on the development set of CoQA for RoBERTa-large. In Figure~\ref{fig:cls-1}, we see that the two histograms follow a similar pattern. The cossim is very high for almost all the cases. This is interesting since it shows that even if a significant chunk is removed from the story, it doesn't affect the CLS embedding in any meaningful way. However, in Figure~\ref{fig:cls-2}, there is a drastic difference between the two histograms. Whereas the histogram for the OT strategy still follows a similar pattern as before, the histogram for IBT shows a significant drop in cossim. This shows that, for most of the cases under IBT, the CLS embedding is heavily affected once the rationale is removed from the story.

\begin{figure}
\centering
\scalebox{.5}{\input{./test-3.tex}}
\caption{RoBERTa-large: Histogram plot of cosine similarity between CLS embedding for OS and TS-R for the two training strategies (OT on left and IBT on right).}
\label{fig:cls-2}
\end{figure}

\begin{figure}
\centering
\scalebox{.5}{\input{./test.tex}}
\caption{RoBERTa-large: Histogram plot of cosine similarity between common tokens of OS and TS for the two training strategies (OT on left and IBT on right).}
\label{fig:cos-1}
\end{figure}

\begin{figure}
\centering
\scalebox{.5}{\input{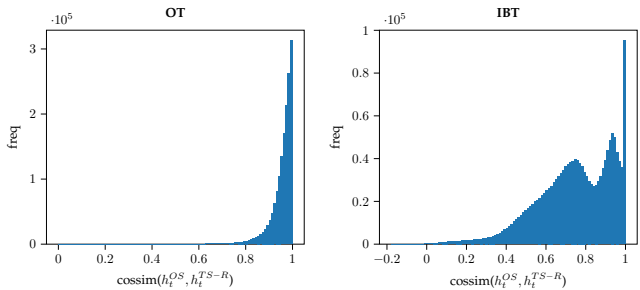}}
\caption{RoBERTa-large: Histogram plot of cosine similarity between common tokens of OS and TS-R under two training strategies (OT on left and IBT on right).}
\label{fig:cos-2}
\end{figure}
This effect is not only local to the CLS token but rather is observed for all the input tokens. To show this, we look at the cosine similarity of common tokens of OS and TS under the two training strategies, and similarly, the cosine similarity of common tokens of OS and TS-R under the two training strategies. Figures~\ref{fig:cos-1} and~\ref{fig:cos-2} show the corresponding histogram for RoBERTa-large. Here also, we can see that the cosine similarity of common tokens in OS and TS is very high for both training strategies.  Once again, the model's representation of the common words doesn't seem to be affected by the removal of large parts of the textual context; this indicates either that the model finds the larger context irrelevant to the task or it might not be capable of encoding long distance contextual information for this task.

For common tokens in OS and TS-R, however, there is a stark contrast between the two training strategies. For OT, the cosine similarity of common words still remain high but for IBT, the cosine similarity drops by a large margin. This shows that, under IBT, the embeddings of the input tokens are more contextualized with respect to the rationale. Due to this, under IBT, the word embeddings get significantly altered once the rationale is removed from the story. Similar to RoBERTa-large, other models also exhibit similar pattern of cosine similarity for CLS and common tokens, as shown in Tables~\ref{tab:cos-cls1} and~\ref{tab:cos-com1} for CoQA, and Table~\ref{tab:cos-hotpot} for HotpotQA. From Table~\ref{tab:cos-cls1}, we can see that, for all the models, cossim($h_{CLS}^{OS},h_{CLS}^{TS-R}$) for IBT is much lower than the corresponding cosine similarity for OT; whereas cossim($h_{CLS}^{OS},h_{CLS}^{TS}$) is similar for both the strategies. Similarly, Table~\ref{tab:cos-com1} shows that, for all the models, cossim($h_{t}^{OS},h_{t}^{TS-R}$) for IBT is much lower than the corresponding cosine similarity for OT; whereas cossim($h_{t}^{OS},h_{t}^{TS}$) is similar for both the strategies. For HotpotQA, Table~\ref{tab:cos-hotpot} shows that IBT has lower cossim($h_{CLS}^{OS},h_{CLS}^{OS-R}$) and cossim($h_{t}^{OS},h_{t}^{OS-R}$) for all the models in comparison to OT.

\begin{table}[t]
\scriptsize
\centering
\begin{tabular}{|c|c|c|c|c|}
\hline
\multirow{3}{*}{Model} & \multicolumn{2}{c|}{OT} & \multicolumn{2}{c|}{IBT}\\ 
\cline{2-5} 
& \multirow{2}{*}{cossim($h_{CLS}^{OS},h_{CLS}^{TS}$)} & \multirow{2}{*}{cossim($h_{CLS}^{OS},h_{CLS}^{TS-R}$)} &\multirow{2}{*}{cossim($h_{CLS}^{OS},h_{CLS}^{TS}$)}& \multirow{2}{*}{cossim($h_{CLS}^{OS},h_{CLS}^{TS-R}$)}\\ & & & &\\
\hline
BERT-base    & $0.99 \pm 0.02$ & $0.97 \pm 0.03$ & $0.96 \pm 0.07$& $0.33 \pm 0.34$\\
\hline
BERT-large    & $0.99 \pm 0.02$ & $ 0.98\pm 0.04$ & $0.95 \pm 0.12$& $0.42 \pm 0.31$\\
\hline
RoBERTa-base    & $0.96 \pm 0.04$ & $0.92 \pm 0.08$ & $0.94 \pm 0.06$& $0.55 \pm 0.22$\\
\hline
RoBERTa-large    & $0.97 \pm 0.06$ & $0.88 \pm 0.10$ & $0.95 \pm 0.07$& $0.47 \pm 0.21$\\
\hline
XLNet-base    & $0.99 \pm 0.03$ & $0.95 \pm 0.09$ & $0.97 \pm 0.06$& $0.27 \pm 0.33$\\
\hline
XLNet-large    & $0.98 \pm 0.04$ & $0.89 \pm 0.15$ & $0.98 \pm 0.05$& $0.53 \pm 0.21$\\
\hline
\end{tabular}
\caption{Cosine similarity (mean$\pm$ std) of CLS embeddings for the two strategies.}
\label{tab:cos-cls1}
\end{table}

\begin{table}[t]
\scriptsize
\centering
\begin{tabular}{|c|c|c|c|c|}
\hline
\multirow{3}{*}{Model} & \multicolumn{2}{c|}{OT} & \multicolumn{2}{c|}{IBT}\\ 
\cline{2-5} 
& \multirow{2}{*}{cossim($h_{t}^{OS},h_{t}^{TS}$)} & \multirow{2}{*}{cossim($h_{t}^{OS},h_{t}^{TS-R}$)} &\multirow{2}{*}{cossim($h_{t}^{OS},h_{t}^{TS}$)}& \multirow{2}{*}{cossim($h_{t}^{OS},h_{t}^{TS-R}$)}\\ & & & &\\
\hline
BERT-base    & $0.99 \pm 0.04$ & $0.94 \pm 0.07$ & $0.96 \pm 0.06$& $0.66 \pm 0.22$\\
\hline
BERT-large    & $0.99 \pm 0.04$ & $0.94 \pm 0.06$ & $0.98 \pm 0.04$& $0.71 \pm 0.21$\\
\hline
RoBERTa-base    & $0.99 \pm 0.04$ & $0.95 \pm 0.06$ & $0.97 \pm 0.05$& $0.75 \pm 0.20$\\
\hline
RoBERTa-large    & $0.99 \pm 0.03$ & $0.95 \pm 0.06$ & $0.98 \pm 0.04$& $0.74 \pm 0.19$\\
\hline
XLNet-base    & $0.96 \pm 0.08$ & $0.90 \pm 0.13$ & $0.96 \pm 0.08$& $0.57 \pm 0.40$\\
\hline
XLNet-large    & $0.94 \pm 0.14$ & $0.86 \pm 0.24$ & $0.94 \pm 0.15$& $0.52 \pm 0.44$\\
\hline
\end{tabular}
\caption{Cosine similarity (mean$\pm$ std) of common tokens for the two strategies.}
\label{tab:cos-com1}
\end{table}

\begin{table}
\scriptsize
\centering
\begin{tabular}{|c|c|c|c|c|}
\hline
\multirow{3}{*}{Model} & \multicolumn{2}{c|}{OT} & \multicolumn{2}{c|}{IBT}\\ 
\cline{2-5} 
& \multirow{2}{*}{cossim($h_{CLS}^{OS},h_{CLS}^{OS-R}$)} & \multirow{2}{*}{cossim($h_{t}^{OS},h_{t}^{OS-R}$)} &\multirow{2}{*}{cossim($h_{CLS}^{OS},h_{CLS}^{OS-R}$)}& \multirow{2}{*}{cossim($h_{t}^{OS},h_{t}^{OS-R}$)}\\ & & & &\\
\hline
BERT-base    & $0.86 \pm 0.12$ & $0.85 \pm 0.15$ & $-0.03 \pm 0.20$& $0.64 \pm 0.20$\\
\hline
BERT-large    & $0.53 \pm 0.42$ & $0.75 \pm 0.21$ & $0.17 \pm 0.14$& $0.55 \pm 0.18$\\
\hline
RoBERTa-base    & $0.91 \pm 0.07$ & $0.84 \pm 0.13$ & $0.32 \pm 0.22$& $0.59 \pm 0.18$\\
\hline
RoBERTa-large    & $0.92 \pm 0.11$ & $0.79 \pm 0.20$ & $0.40 \pm 0.17$& $0.54 \pm 0.16$\\
\hline
XLNet-base    & $0.96 \pm 0.05$ & $0.92 \pm 0.15$ & $0.00 \pm 0.23$& $0.87 \pm 0.20$\\
\hline
XLNet-large    & $0.81 \pm 0.25$ & $0.91 \pm 0.16$ & $-0.01 \pm 0.18$& $0.77 \pm 0.29$\\
\hline
\end{tabular}
\caption{Cosine similarity (mean$\pm$ std) of CLS and common token embeddings for HotpotQA}
\label{tab:cos-hotpot}
\end{table}
From a more conceptual perspective, the sensitivity to the rationale in IBT suggests that IBT is providing the kind of instances needed to confirm the counterfactual, {\em were the rationale not present, the model would not answer as it does when the rationale is present}.  Thus, at a macro level, attention based models can locate spans of text crucial to determining semantic content through particular forms of training.

\section{Negation Intervention}\label{sec:neg}

In this section, we detail our experiments on negation intervention. Negation intervention investigates possible causal dependencies of a model's inferences based on logical structure, in particular the scope of negation operators.  As we said in Section \ref{sec:sem}, the idea behind negation intervention is to alter a text with an intervention $n$ such that $T, Q \mmodels \psi \ \mbox{iff }  n(T), Q \mmodels \neg \psi$.

For negation intervention, we randomly sampled $275$ yes-no questions for CoQA. We appropriately modified the rationale in the truncated story (i.e., TS) for these samples in order to switch the answer from yes to no and vice-versa. The development set of HotpotQA contained $458$ yes-no questions. Out of these, we randomly sampled $159$ questions. Almost all the questions followed the format \emph{``Were both X and Y Z?"} (e.g., ``Were Scott Derrickson and Ed Wood of the same nationality?"). For these samples, we modify the rationale in the original story (i.e., OS) in order to switch the answer. Table~\ref{tab:neg} shows the effect of negation intervention for the model trained under OT for the two datasets. In the table, Org-Acc refers to accuracy of the model on the original sample, Mod-Acc refers to accuracy of the model post negation intervention (i.e., with respect to the modified ground truth answer), and Comb-Acc refers to the percentage of cases where the model answered correctly for both original and modified sample. Table~\ref{tab:neg} shows a $\sim 20\%$ drop in accuracy for CoQA for all the models when we compare Org-Acc and Mod-Acc. For HotpotQA, the drop in accuracy is even higher for BERT, and XLNet. This significant drop highlights the inability of the models to handle negation intervention. The low Comb-Acc scores of the models across the two datasets further highlight this fact.  Switching to the IBT regime provided no significant difference. This indicates that another type of training will be needed for these models to take into systematic account the semantic contributions of negation.  

A natural option is to train over negated examples and non negated examples.  \namecite{kassner-schutze-2020-negated}
performs such an experiment and concludes that transformer style models do not learn the meaning of negation.  And \namecite{hosseini-etal-2021-understanding} provides a particular training regime that seems to improve language models' performance on the dataset of negated examples introduced by \namecite{kassner-schutze-2020-negated}. 
\begin{table}[t]
\centering
\small
\begin{tabular}{|c|c|c|c||c|c|c|}
\hline
\multirow{2}{*}{Model} & \multicolumn{3}{c||}{CoQA} & \multicolumn{3}{c|}{HotpotQA} \\ \cline{2-7}
& Org-Acc & Mod-Acc & Comb-Acc & Org-Acc & Mod-Acc & Comb-Acc\\
\hline
BERT-base    & 78.2 & 58.9 & 41.5 & 68.6 & 39.0 & 8.8\\
\hline
BERT-large    & 84.7 & 65.1 & 52.0  & 71.1 & 39.0 & 10.1 \\
\hline
RoBERTa-base    & 81.8 & 61.8 & 47.6 & 61.6 & 45.9 & 9.4\\
\hline
RoBERTa-large    & 94.2 & 72.7 & 67.3 & 84.9 & 62.3 & 49.1\\ 
\hline
XLNet-base    & 85.1 & 64.7 & 52.0 & 69.8 & 37.1 & 6.9\\
\hline
XLNet-large    & 90.2 & 68.7 & 59.6& 84.3 & 47.8 & 33.3\\ 
\hline
\end{tabular}
\caption{Effect of Negation intervention on different models.}
\label{tab:neg}
\end{table}

Nevertheless, while \namecite{kassner-schutze-2020-negated}'s conclusion is compatible with our findings, we are not sanguine that \namecite{hosseini-etal-2021-understanding}'s training regime will improve model performance on the operation of negation intervention. 
The interventions we needed to make to induce the appropriate shifts in answers often depended on quite important shifts in material.   Simple insertions of negation often seemed to disrupt the coherence and flow of the text; these disruptions could provide superficial clues for shifting the model's behavior in a task.  To give an example, here is a rationale from one of the stories in the CoQA dataset:

\ex. \label{law}
A law enforcement official told The Dallas Morning News that a door was apparently kicked in

Given the question, {\em Was the house broken into?} (the original answer was {\em yes}), we changed the rationale to:

\ex. \label{law'}
A law enforcement official told The Dallas Morning News that a door was open, leaving the possibility that the killers had been invited in

\noindent
to get a negative answer to the question.

In this intervention, we didn't insert a negation but rather changed the wording to get a text inconsistent with the original answer. More generally, in only 72  out of 275 cases (26 \%),  we added or removed "no/not". And within these 72 cases, only around 25 cases featured the simple addition/removal of "no/not"---e.g. the replacement of {\em six corn plants} to {\em but no corn plants}. In the rest of the cases, although we added/removed "no/not", we made more substantive changes to the story.  Here is one example, where the question was, {\em did the wife console the boy?} The original rationale was as follows:

\ex. \label{meyers}
"Robert Meyers said his wife tried to help Nowsch.  "My wife spent countless hours at that park consoling this boy," he said.

We changed this to:
\ex. \label{meyers'}
Robert Meyers said his wife did not try to help Nowsch. "My wife spent countless hours at that park tormenting this boy," he said.

\begin{table}[t]
\centering
\small
\begin{tabular}{|c|c|c|c|c|}
\hline
Model & Dataset & Org-Acc & Mod-Acc & Comb-Acc \\
\hline
\multirow{2}{*}{text-davinci-002}    & CoQA & 88.0 & 61.1 & 53.5 \\
                   & HotpotQA & 79.2 & 49.7 & 31.4 \\
\hline
\multirow{2}{*}{text-davinci-003}   & CoQA & 94.2 & 61.5 & 56.7 \\
                   & HotpotQA  & 90.0 & 59.7 & 50.9 \\ 
\hline
\end{tabular}
\caption{Effect of Negation intervention on the two InstructGPT models.}
\label{tab:gpt3-negintnew}
\end{table}
For the other 203 out of 275 cases (74\%), there were lexical changes and substantial changes to the rationale to preserve stylistic consistency.  

Thus the simple addition/removal of "no/not"  cases numbered around 25 cases ($\sim 10\%$). In general, the models were able to switch their answers on such cases. Out of 73 cases, where Roberta-large answered the question for the original story correctly and didn't switch the answer post negation intervention, there are only $6$ trivial cases. For HotpotQA, given the format of the question, simply adding/removing "no/not" is insufficient. For all the $159$ cases, switching the answer required manipulating the entities present in the story (e.g., changing \emph{American} to \emph{Australian}). 

Similar to deletion intervention, we also test InstructGPT on our negation intervention dataset. Table~\ref{tab:gpt3-negintnew} shows the results. From the tables, we can see that both \emph{text-davinci-002} and \emph{text-davinci-003} suffer a $\sim 20\%$ drop in accuracy for CoQA and $\sim 30\%$ drop in accuracy for HotpotQA. Thus, similar to other models, InstructGPT fails to respond well to negation intervention. The inferences involved in negation intervention are thus quite complex and go beyond the recognition of a simple negation.  A mastery of such inferences would indicate a mastery not only of negation but of inconsistency, which would be a considerable achievement for a machine learned model.  So simply alerting the model to the presence of negation will not suffice to guarantee reasoning ability with negation. The alternative is to create a corpus with many more negation intervention examples.  However, it was difficult to construct the requisite data so as to meet our view of negation intervention since fine-tuning requires a lot more examples. We provide several examples of our negation intervention dataset for CoQA and HotpotQA in Section~\ref{sec:appendix}.

\section{Predicate\textendash Argument Structure}\label{sec:pred}

In this section, we study whether the models stay faithful to simple cases of predicate\textendash argument structure.  As we already mentioned in the introduction, we propose two types of experiments.  In the first simple experiment, we ask a question $Q$ about the properties of objects in a text $T$.  Given an answer $\psi$ such that $T, Q 
\mmodels \psi$, we expect that for a semantically faithful model $M_T$
that $M_T, Q \mmodels \psi$.

The second set of experiments is more involved.  Formally, it involves the following set up.  Given:
\begin{itemize}
    \item two questions, $Q, Q'$, 
    \item $T \models Q \leftrightarrow Q'$
\end{itemize}
we should have 
$$  T, Q \mmodels \psi \ \mathit{iff} \ T, Q' \mmodels \psi$$

To test for semantic faithfulness in these contexts, we devised synthetic, textual data for these experiments.  We used five different schemas:

\begin{enumerate}
    \item 
The \textit{col1} car was standing in front of a \textit{col2} house.
    \item 
They played with a \textit{col1} ball and \textit{col2} bat.
    \item 
The man was wearing a \textit{col1} shirt and a \textit{col2} jacket.
    \item 
The house had a \textit{col1} window and a \textit{col2} door.
    \item 
A \textit{col1} glass was placed on a \textit{col2} table.
\end{enumerate}

where \textit{col1} and \textit{col2} denote two distinct colors. Using these $5$ schemas and different color combinations, we constructed a dataset of $130$ stories. For each story, we have $4$ questions. (2 ``yes" and 2 ``no" questions). As an example, for the story, ``The blue car was standing in front of a red house.", the $2$ ``yes" questions are ``Was the car blue?" and ``Was the house red?"; and 2 ``no" questions are ``Was the car red?" and ``Was the house blue?". Thus, we have a total of $520$ questions. The Org-Acc in Table~\ref{tab:paraphrase} shows the accuracy of the models trained on CoQA on these questions and indicates a huge variance in accuracy across the models. We observed that BERT-base predicted \emph{no} for all questions, and RoBERTa-base predicted \emph{no} for most of the questions, while XLNet-base mostly predicted ``yes''.  For the large models, RoBERTa-large and BERT-large achieved very high accuracies.  We note, however, that this dataset is very simple.

\begin{table}[t]
\centering
\footnotesize
\begin{tabular}{|c|c|c|c|}
\hline
Model & Org-Acc & Mod-Acc & New-Acc\\
\hline
BERT-base    & 50.0 (100.0) & 69.4 (59.8) & 50.0 (100.0) \\
\hline
BERT-large    & 95.2 (51.0) & 77.3 (27.3) & 98.1 (51.5)  \\
\hline
RoBERTa-base    & 51.0 (99.0) & 70.0 (78.5) & 50.6 (99.4) \\
\hline
RoBERTa-large    & 99.4 (49.4) & 95.0 (45.0) & 99.8 (49.9) \\ 
\hline
XLNet-base    & 50.6 (6.0) & 50.8 (0.7)  & 59.0 (13.3) \\
\hline
XLNet-large    & 75.2 (74.8) & 79.8 (36.3) & 78.7 (68.7) \\ \hline
\end{tabular}
\caption{Effect of question paraphrasing on different models trained on CoQA. Org-Acc, and Mod-Acc denote accuracy on original and modified paraphrased question respectively. The number in bracket denote percentage of cases where the model predicted ``no" as the answer. New-Acc refers to models' performance when ``there" is replaced by ``the" in the modified question.}
\label{tab:paraphrase}
\end{table}

These results indicate that the small models really didn't do much better than chance in answering our yes/ no questions; hence either they didn't capture of the predicate\textendash argument structure of the sentences, or they could not use that information to reason to an answer to our questions.  They failed on the most basic level. The large models fared much better, but this in itself didn't suffice to determine a causal link between the predicate\textendash argument information and the inference to the answers.



Probing further, we then examined how the models fared under semantically equivalent questions.  $Q'$ is semantically the same as ($\equiv$) $Q$ given context $C$ iff they have the same answer sets in $C$ \cite{bennett:1979,karttunen:1977,groenendijk:2003}. In our situation, the context is given by the text $T$.  Thus, we have $T \models Q \leftrightarrow Q'$ and $T, Q \mmodels \psi \ \mathit{iff} \ T, Q' \mmodels \psi$. If $M_T$ is semantically faithful and $T \models Q \leftrightarrow Q'$, then we should have $M_T,Q \mmodels \psi$ iff $M_T,Q' \mmodels \psi$.  To construct semantically equivalent questions, we paraphrased the initial question in our dataset, i.e., {\em ``Was the car red?"} to {\em ``Was there a red car?"}. This resulted in new set of $520$ questions for the $130$ stories. The Mod-Acc in Table~\ref{tab:paraphrase} shows the accuracy of the models trained on CoQA on the modified questions. Apart from XLNet-base which predicted ``yes" for most of the modified questions and RoBERTa-large which retains it high accuracy, all the other models behave very differently from before. The accuracy of BERT-large drops drastically on these very simple questions, while BERT-base and RoBERTa-base perform significantly better on modified questions as they no longer mostly predict  ``no".  For XLNet-large, while the two accuracies are similar, the model goes from predicting mostly ``no" to mostly ``yes". This contrast in behavior, as shown in Table~\ref{tab:paraphrase}, indicates that these models are unstable and lack semantic faithfulness on this task. There could be two possible reasons for this contrast, (i) different ordering of predicate and argument (``ball \textit{col1}" vs. ``\textit{col1} ball"), (ii) different surface level words (``the" vs. ``there"). We found the latter to be the case. To show this, we replace ``there" with ``the" in the modified question (i.e., ``Was there a \textit{col1} ball?" $\rightarrow$ ``Was the a \textit{col1} ball?"). The New-Acc in Table~\ref{tab:paraphrase} shows the models' performance on this new question. We can see that the New-Acc is consistent with Org-Acc both in terms of accuracy and percentage of \emph{no} predictions. This shows that the models are extremely sensitive to semantically unimportant words. For the models trained on HotpotQA, for both the question types (org, and mod), we found that they either failed to recognize the question type as yes/no, or predicted ``no". This is likely due to the fact that the yes/no questions in HotpotQA follow a fixed format, as discussed in Section~\ref{sec:neg}, which is different from the one being used for this experiment. 

\begin{figure}
\centering
\includegraphics[width=0.85\linewidth]{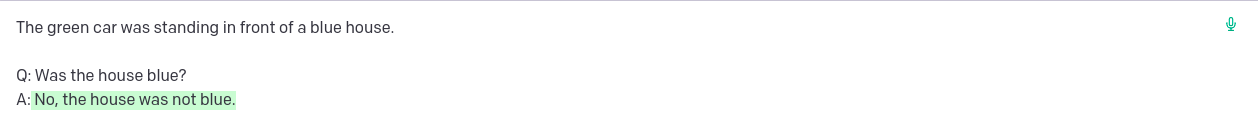}\\
\includegraphics[width=0.85\linewidth]{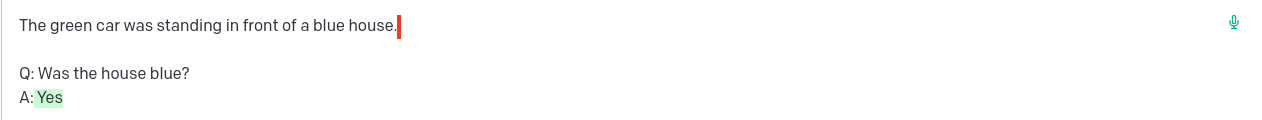}
\caption{The \emph{text-davinci-002} model predicts correctly when an extra space (shown in red) is added.}
\label{fig:gpt}
\end{figure}

Finally, we evaluate the performance of InstructGPT models. On the overall dataset of $1040$ questions ($520$ org questions and $520$ mod questions), \emph{text-davinci-002} achieved an accuracy of $96.7\%$ (i.e. total of $34$ failure cases) with Org-Acc of $99.6\%$ and Mod-Acc of $93.8\%$. Interestingly, all the $34$ failure cases in the original predicate\textendash argument dataset were ``yes" questions. For such cases, in many instances, we observed that adding an extra space to the prompt reverses the model's prediction. One such example is shown in Figure~\ref{fig:gpt}. As for \emph{text-davinci-003}, the model achieved perfect accuracy. Unlike \emph{text-davinci-002}, we found that \emph{text-davinci-003} is stable in its prediction with regards to extra spaces in the prompt. However, there were $14$ cases where \emph{text-davinci-003} predicted ``not necessarily" instead of ``no". The question in all these cases was of the form ``Was there a \textit{col2} car?" Note that we had $26$ cases with this question format and the model predicted ``no" for the remaining $12$ cases. This showcases instability in model's prediction for two very similar input  prompts.

We also tested the model's sensitivity to a contrastive set of examples in which we inserted negations in our predicate\textendash argument dataset sentences (e.g. ``The blue car was standing in front of a house that was not red." for the question ``Was the house red?").  In contrast to its performance on negation intervention, the InstructGPT models achieved perfect accuracy on such negated examples.  This further demonstrates that negation intervention is different from the tasks given by \namecite{naik-etal-2018-stress, kassner-schutze-2020-negated, hossain-etal-2020-analysis, hosseini-etal-2021-understanding}.

\section{Discussion}\label{sec:disc}

In conclusion, concerning our findings about predicate\textendash argument structure and logical structure more generally, we address three points.


 1) Larger Transformer-based models have shown to generally perform better than their smaller variants~\cite{roberts-etal-2020-much, liu-etal-2019-linguistic}. However, some exceptions to this trend have also been observed~\cite{zhong-etal-2021-larger}. Our experiments show that with respect to the notion of semantic faithfulness, in general sensitivity to semantic structure and content, larger models fare better in predicate\textendash argument experiments but not in our negation and deletion intervention experiments. For deletion intervention, they are mostly worse-off than smaller models. Sections~\ref{sec:del} and~\ref{sec:neg} show that InstructGPT also fails to tackle the two interventions in an efficient manner. For predicate\textendash argument structure, the fact that the difference in models' behavior to two question types arises from irrelevant surface-level words is a big drawback since there are multiple ways to paraphrase a question. A possible way to tackle this issue is to translate the question into a logical form and train the model so that the contextualised embedding of the argument  is more sensitive to its corresponding predicate. For example, the contextualised embedding of ``car" is more sensitive to ``blue" for the story ``The blue car was standing in front of a red house." The sensitivity of a word $w$ on another word $w'$ can quantitatively be measured as the change in the contextualised embedding of the word $w$ when the word $w'$ is removed from its context. Higher the change, higher the sensitivity. This sensitivity can then be used as a cue by the model to arrive at a prediction. We plan to explore this approach in future.

2) Is prompting really superior to fine tuning? 
 Our prompting experiment with InstructGPT allowed us to get results without fine-tuning.  This is essentially zero shot learning since no input-output pairs are provided in the prompt. However, for deletion and negation intervention, we observed that InstructGPT models do not present an advancement over other Transformer-based models with respect to behavior post these interventions.  Moreover, like~\namecite{jiang:etal:2020,liu:etal:2021,shin:etal:2021constrained}, we have found \emph{text-davinci-002} to be extremely sensitive to what should be and intuitively is irrelevant information in the prompt. With regard to semantic faithfulness on predicate\textendash argument structure, this shows an astonishing lack of robustness to totally irrelevant material, even if \emph{text-davinci-002} scores very well on this dataset.  This brittleness is telling; a semantically faithful model that exploits semantic structure to answer questions about which objects have which properties should not be sensitive to formatting changes in the prompt.  This indicates to us that even if predicate\textendash argument structure questions are answered correctly, \emph{text-davinci-002} is not using that information as it should. \emph{text-davinci-003} is stable to such insignificant changes in the prompt. However, the model still shows instability in its predictions for two very similar prompts as highlighted earlier.  Once again, we have our doubts that the right information, i.e. semantic structure, is being leveraged for the answer; if it were, \emph{text-davinci-003} would answer in the same way for all the questions with "no" answers.  
 
3) Extending semantic faithfulness beyond the question answering tasks in NLP. The definition of semantic faithfulness in Section~\ref{sec:sem} is geared to testing the semantic knowledge of LMs in question answering tasks.  Question answering can take many forms and is a natural way to investigate many forms of inference or exploitations of semantic and logical structure.  It also underlies many real-world NLP applications, like chatbots, virtual assistants and web searches \cite{liang2022holistic}.  Semantic faithfulness can be extended to probe for a model's inferences concerning artificial languages like first order logic or any other formal language or programming language for which there is a well defined notion of semantic consequence ($\models$).  In such cases, the role of the ``text'' in semantic faithfulness would be played by a set of premises, a logical or mathematical theory, or code for an algorithm or procedure.  Similar experiments of deletion or negation intervention could in principle be performed in these settings, which opens up a novel way of investigating LM models' performance on tasks like code generation \cite{chen:etal:2021,sarsa:etal:2022}.  Alternatively, as suggested by~\namecite{shin:etal:2021few}, exploiting formal logical forms may help with semantic faithfulness.


\section{Conclusion and Future Work}\label{sec:conc}

We have studied the semantic faithfulness of Transformer-based language models for two intervention strategies, deletion intervention, and negation intervention, and with respect to their responses to simple, semantically equivalent questions. Despite high performance on the original CoQA and HotpotQA, the models exhibited very low sensitivity to deletion intervention  and suffered a significant drop in accuracy for negation intervention. They also exhibited unreliable and unstable behavior with respect to semantically equivalent questions (Q$\equiv$).  
Our simple intervention-based training (IBT) strategy made the contextualized embeddings more sensitive to the rationale and corrected the models' erroneous reasoning in the case of deletion intervention. 

Our paper has exposed flaws in popular language models.  In general, we have shown that even large models are not guaranteed to respect semantic faithfulness. This likely indicates that the models rely on superficial cues for answering questions about a given input text. While IBT is successful at remedying models' lack of attention to logical structure in cases of deletion intervention, it doesn't generalize well to the other experimental setups we have discussed. We do not have easy fixes for negation interventions or for the inferences involving predicate\textendash argument structure. This is because it is difficult to generate enough data through negation intervention to retrain the model in the way we did for deletion intervention.  Automating the process of negation intervention while preserving a text's discourse coherence and particular style remains a challenge. In addition, our investigations concerning predicate\textendash argument structure and responses to semantically equivalent questions have pointed to a serious failing. We plan to explore the proposed approach in Section~\ref{sec:disc} to tackle this issue. Also, this work focused on a specific setting of semantic faithfulness where $\phi = \psi$ (refer to Equation~\ref{sem}). As part of future work, we plan to study the setting where multiple answers are possible. 

Deletion and negation intervention modify the semantics of the input text. On the other hand, question equivalence experiments in this work preserves the semantics of the input. We plan to analyse the models' behavior to other possible semantic preserving interventions. A general solution to the problem of semantic unfaithfulness is something we have not provided in this paper.  However, we believe that key to solving this problem is a full scale integration of semantic structure without loss of inferential power in the transformer based language models, something we plan to show in future work.

\section{Appendix}\label{sec:appendix}
\setcounter{table}{0}
\renewcommand{\thetable}{A. \arabic{table}}
Tables~\ref{tab:del-inter-ex1},~\ref{tab:del-inter-ex2},~\ref{tab:del-inter-ex3},~\ref{tab:del-inter-ex4} show examples of deletion intervention for models trained under OT for CoQA and HotpotQA. Similarly, Tables~\ref{tab:neg-inter-ex1},~\ref{tab:neg-inter-ex2} show examples of negation intervention for CoQA and HotpotQA respectively. In all the examples presented here, the models' prediction remains unchanged post intervention showing that the models rely on superficial cues for predicting an answer to a given question.

\begin{table}
    \scriptsize
    \centering
    \begin{subtable}[t]{1\linewidth}
    \centering
    \(\begin{tabular}{l|l} 
    \hline
        \multirow{3}{*}{\textbf{TS}} &  My doorbell rings. On the step, I find the elderly Chinese lady, \\ & small and slight, holding the hand of a little boy. \textbf{In her other} \\& \textbf{hand, she holds a paper carrier bag.} \\ \hline
        \multirow{2}{*}{\textbf{TS-R}} &  My doorbell rings. On the step, I find the elderly Chinese lady, \\ & small and slight, holding the hand of a little boy. paper carrier bag.\\ \hline
          \textbf{Conversation }& Who is at the door? An elderly Chinese lady and a little boy \\ \textbf{History }& Is she carrying something? yes \\ \hline
          \textbf{Question}& What? \\ \hline
          \textbf{Prediction} & a paper carrier bag \\ \hline
    \end{tabular}\)
    \caption{BERT-base}
    \medskip
    
    \(\begin{tabular}{l|l}
    \hline
        \multirow{5}{*}{\textbf{TS}} &  OCLC, currently incorporated as OCLC Online Computer Library \\ & Center, Incorporated, is an American nonprofit cooperative  \\ & organization ``dedicated to the public purposes of furthering access \\ & to the world's information and reducing information costs". \\ & \textbf{It was founded in 1967 as the Ohio College Library Center.}\\ \hline
        \multirow{4}{*}{\textbf{TS-R}} &  OCLC, currently incorporated as OCLC Online Computer Library \\ & Center, Incorporated, is an American nonprofit cooperative  \\ & organization ``dedicated to the public purposes of furthering access \\ & to the world's information and reducing information costs". 1967.\\ \hline
          \textbf{Conversation }& What is the main topic? OCLC \\ \textbf{History }& What does it stand for? Online Computer Library Center \\ \hline
          \textbf{Question}& When did it begin? \\ \hline
          \textbf{Prediction} & 1967 \\ \hline
    \end{tabular}\)
    \caption{BERT-large}
    \medskip
    \(\begin{tabular}{l|l}
    \hline
        \multirow{5}{*}{\textbf{TS}} &  Chapter XVIII ``The Hound Restored" On the third day after his  \\& arrival at the camp Archie received orders to prepare to start with \\& the hound, with the earl and a large party of men-at-arms, in search \\&  of Bruce. \textbf{A traitor had just come in and told them where Bruce} \\&  \textbf{had slept the night before.}\\ \hline
        \multirow{4}{*}{\textbf{TS-R}} &  Chapter XVIII ``The Hound Restored" On the third day after his\\&  arrival at the camp Archie received orders to prepare to start with \\& the hound, with the earl and a large party of men-at-arms, in search \\&  of Bruce. A traitor.\\ \hline
         \textbf{Conversation} & What was he told to start to do? Search for Bruce \\{\textbf{History}} & With what? with the hound, with the earl and a large party of \\ & men-at-arms \\ \hline
          \textbf{Question}& Who gave them information about Bruce? \\ \hline
          \textbf{Prediction} & A traitor \\ \hline
    \end{tabular}\)
    \caption{RoBERTa-base}
    \medskip
    \(\begin{tabular}{l|l}
    \hline
        \multirow{4}{*}{\textbf{TS}} &  Can you imagine keeping an alien dog as a pet? This is what \\&  happens in CJ7 [...] When Ti falls off a building and dies, CJ7 saves  \\& his life. Because the dog loses all its power, it becomes a doll. \textbf{But}  \\& \textbf{Dicky still wears the dog around his neck.}\\ \hline
        \multirow{4}{*}{\textbf{TS-R}} &  Can you imagine keeping an alien dog as a pet? This is what \\&  happens in CJ7 [...] When Ti falls off a building and dies, CJ7 saves  \\& his life. Because the dog loses all its power, it becomes a doll. \\&  around his neck.\\ \hline
          \textbf{Conversation }& What did he become? a doll \\ \textbf{History }& True or False: the boy loses the doll? False \\ \hline
          \textbf{Question}& Where does he keep it, then? \\ \hline
          \textbf{Prediction} & around his neck \\ \hline
    \end{tabular}\)
    \caption{RoBERTa-large}
    \end{subtable}
    \caption{Deletion Intervention Examples from CoQA dataset. The rationale is marked in bold in TS. The models predict the same answer for TS and TS-R.}
    \label{tab:del-inter-ex1}
\end{table}

\begin{table}
    \scriptsize
    \centering
    \begin{subtable}[t]{1\linewidth}
    \centering
    \(\begin{tabular}{l|l} 
    \hline
        \multirow{5}{*}{\textbf{TS}} &  Dhaka is the capital and largest city of Bangladesh. [...] At the \\ &  height of its medieval glory, Dhaka was regarded as one of  \\& the wealthiest and most prosperous cities in the world. \textbf{It } \\& \textbf{served as the capital of the Bengal  province of the Mughal Empire} \\&    \textbf{twice (1608–39 and 1660–1704).}\\ \hline
        \multirow{4}{*}{\textbf{TS-R+Aug}} & Dhaka is the capital and largest city of Bangladesh. [...] At the \\ &  height of its medieval glory, Dhaka was regarded as one of \\&  the wealthiest and most prosperous cities in the world. I  had to read \\&  the book twice to fully  understand its theme.\\ \hline
        \textbf{Conversation }& Was it ever one of the wealthiest cities in the world? yes \\ \textbf{History }& and when was that? At the height of its medieval glory \\ \hline
          \textbf{Question}& How many times was it the capital of the Bengal province? \\ \hline
          \textbf{Prediction} & twice \\ \hline
    \end{tabular}\)
    \caption{RoBERTa-base}
    \medskip
    
    \(\begin{tabular}{l|l}
    \hline
        \multirow{2}{*}{\textbf{TS}} &  Andrew waited for his granddaddy to show up. They were going  \\&  fishing. \textbf{His mom had packed them a lunch.}\\ \hline
        \multirow{2}{*}{\textbf{TS-R+Aug}} &  Andrew waited for his granddaddy to show up. They were going  \\&  fishing. I packed them a lunch for their long road trip.\\ \hline
          \textbf{Conversation }& What was Andrew waiting for? His granddaddy \\ \textbf{History }& Why? They were going fishing \\ \hline
          \textbf{Question}& What did his mom do?\\ \hline
          \textbf{Prediction} & packed them a lunch \\ \hline
    \end{tabular}\)
    \caption{RoBERTa-large}
    \medskip
    \(\begin{tabular}{l|l}
    \hline
        \multirow{3}{*}{\textbf{TS}} &  ATLANTA, Georgia (CNN) -- In 1989, the warnings were dire.  \\& \textbf{The Spike Lee film ``Do the Right Thing" critics and columnists}  \\& \textbf{said, would provoke violence and disrupt race relations.}\\ \hline
        \multirow{3}{*}{\textbf{TS-R+Aug}} &  ATLANTA, Georgia (CNN) -- In 1989, the warnings were dire.  \\& Spike Lee is a highly acclaimed filmmaker known for his innovative \\& and thought-provoking films.\\ \hline
         \textbf{Conversation} & \multirow{2}{*}{-} \\{\textbf{History}} &  \\ \hline
          \textbf{Question}& Who created Do the Right Thing?\\ \hline
          \textbf{Prediction} & Spike Lee \\ \hline
    \end{tabular}\)
    \caption{XLNet-base}
    \medskip
    \(\begin{tabular}{l|l}
    \hline
        \multirow{4}{*}{\textbf{TS}} &  Once upon a time there was a cute brown puppy. He was a\\&  very happy puppy. His name was Rudy. Rudy had a best \\&  friend. His name was Thomas. Thomas had a nice dad named Rick.\\& \textbf{Thomas and Rudy had been friends for almost a year.} \\ \hline
        \multirow{4}{*}{\textbf{TS-R+Aug}} &  Once upon a time there was a cute brown puppy. He was a\\&  very happy puppy. His name was Rudy. Rudy had a best \\&  friend. His name was Thomas. Thomas had a nice dad named Rick.\\& I haven't seen my family in almost a year due to the pandemic. \\ \hline
          \textbf{Conversation }& \multirow{2}{*}{Who was Rudy's best friend? Thomas} \\ \textbf{History }&  \\ \hline
          \textbf{Question}& How long have they been friends? \\ \hline
          \textbf{Prediction} & almost a year \\ \hline
    \end{tabular}\)
    \caption{XLNet-large}
    \end{subtable}
    \caption{Deletion Intervention Examples from CoQA dataset. The rationale is marked in bold in TS. The models predict the same answer for TS and TS-R+Aug.}
    \label{tab:del-inter-ex2}
\end{table}

\begin{table}
    \scriptsize
    \centering
    \begin{subtable}[t]{1\linewidth}
    \centering
    \(\begin{tabular}{l|l} 
    \hline
        \multirow{7}{*}{\textbf{OS}} &  \textbf{Sergio Pérez Mendoza (born 26 January 1990) also known as ``Checo"} \\& \textbf{Pérez, is a Mexican racing driver, currently driving for Force India.} There  \\&  have been  six Formula One drivers from Mexico who have taken  part in    \\&  races since the championship began in 1950. \textbf{Pedro Rodríguez is  the most}  \\&  \textbf{successful Mexican driver being the only one to have won a grand prix.} \\&  \textbf{Sergio Pérez, the only other Mexican to finish on the podium,  currently}  \\& \textbf{races with Sahara Force India F1 Team.} \\ \hline
        \multirow{2}{*}{\textbf{OS-R}} &  There have been  six Formula One drivers from Mexico who have taken      \\& part in races since the championship began in 1950. Pedro Rodríguez \\ \hline
          \multirow{2}{*}{\textbf{Question}} & Which other Mexican Formula One race car driver has held the \\ & podium besides the Force India driver born in 1990? \\ \hline
          \textbf{Prediction} & Pedro Rodríguez  \\ \hline
    \end{tabular}\)
    \caption{BERT-base}
    \medskip
    
    \(\begin{tabular}{l|l}
    \hline
        \multirow{9}{*}{\textbf{OS}} &  \textbf{The Manchester Terrier is a breed of dog of the smooth-haired}\\&  \textbf{terrier type. The Scotch Collie is a landrace breed of dog which originated}\\&  \textbf{from the highland regions of Scotland.} The breed consisted of both\\&   the long-haired (now known as Rough) Collie and the short-haired (now \\&   known as Smooth) Collie. It is generally believed to have descended\\&  from a variety of ancient herding dogs, some  dating back to the Roman\\&  occupation, which may have included Roman Cattle Dogs, Native Celtic \\&    Dogs and Viking Herding Spitzes. \textbf{Other ancestors include the Gordon}  \\& \textbf{and Irish Setters.}\\ \hline
        \multirow{6}{*}{\textbf{OS-R}} &  The breed consisted of both the long-haired (now known as Rough) \\& Collie and the short-haired (now  known as Smooth) Collie. It is generally\\& believed to have descended from a variety of ancient herding dogs, some\\&    dating back to the Roman occupation, which may have included\\&   Roman Cattle Dogs, Native Celtic    Dogs and Viking Herding Spitzes. \\& Scotch Collie \\ \hline
           \multirow{2}{*}{\textbf{Question}}& Which dog's ancestors include Gordon and Irish Setters: the Manchester \\& Terrier or the Scotch Collie?\\ \hline
          \textbf{Prediction} & Scotch Collie \\ \hline
    \end{tabular}\)
    \caption{BERT-large}
    \medskip
    \(\begin{tabular}{l|l}
    \hline
        \multirow{9}{*}{\textbf{OS}} &  Carrefour S.A. is a French multinational retailer headquartered in Boulogne\\&  Billancourt, France, in the Hauts-de-Seine Department near Paris. \textbf{It is} \\&  \textbf{one of the largest hypermarket chains in the world (with 1,462 hypermarkets} \\& \textbf{at the end of 2016).} Carrefour operates in more than 30 countries, in\\&    Europe,the Americas, Asia and Africa. Carrefour means ``crossroads" and \\&  ``public square" in French. The company is a component of the Euro Stoxx\\&    50 stock market index. Euromarché (``Euromarket") was a French  hypermarket \\&  chain. The first store opened in 1968 in Saint-Michel-sur-Orge. \textbf{In June} \\&  \textbf{1991, the group was rebought by its rival, Carrefour, for 5,2 billion francs.}\\ \hline
        \multirow{7}{*}{\textbf{OS-R}} &   Carrefour S.A. is a French multinational retailer headquartered in Boulogne\\&  Billancourt, France, in the Hauts-de-Seine Department near Paris. Carrefour \\& operates in more than 30 countries, in    Europe,the Americas, Asia and \\&  Africa. Carrefour means ``crossroads" and  ``public square" in French.  The\\& company is a component of the Euro Stoxx    50 stock market index.\\& Euromarché (``Euromarket") was a French  hypermarket  chain. The first store \\& opened in 1968 in Saint-Michel-sur-Orge. 1,462 \\ \hline
          \multirow{2}{*}{\textbf{Question}}& In 1991 Euromarché was bought by a chain that operated how any \\& hypermarkets at the end of 2016? \\ \hline
          \textbf{Prediction} & 1,462 \\ \hline
    \end{tabular}\)
    \caption{RoBERTa-base}
    \end{subtable}
    \caption{Deletion Intervention Examples from HotpotQA dataset. The rationale is marked in bold in OS. The models predict the same answer for OS and OS-R.}
    \label{tab:del-inter-ex3}
\end{table}

\begin{table}
    \scriptsize
    \centering
    \begin{subtable}[t]{1\linewidth}
    \centering
    \(\begin{tabular}{l|l} 
    \hline
        \multirow{9}{*}{\textbf{OS}} &  \textbf{Marie Magdalene ``Marlene" Dietrich (27 December 1901 – 6 May 1992)} \\&  \textbf{was a German actress  and singer who held both German and American}  \\& \textbf{citizenship.}  Throughout her unusually long career, which spanned from  \\& the 1910s to the 1980s, she maintained popularity by continually reinventing \\&  herself. \textbf{Marlene, also known in Germany as Marlene Dietrich-Porträt} \\& \textbf{eines Mythos, is a 1984 documentary film made by Maximilian Schell about} \\& \textbf{the legendary  film  star Marlene Dietrich.} It was made by Bayerischer \\&  Rundfunk (BR) and OKO-Film and released by Futura Film, Munich and \\& Alive Films (USA). \\ \hline
        \multirow{5}{*}{\textbf{OS-R+Aug}} & Throughout her unusually long career, which spanned from  the 1910s to \\& the 1980s,  she maintained popularity by continually reinventing \\&  herself. It was made by Bayerischer  Rundfunk (BR) and OKO-Film  and \\&  released by Futura Film, Munich and Alive Films (USA). The year 1901 \\& marked the beginning of a new century. \\  \hline
        \multirow{2}{*}{\textbf{Question}} & The 1984 film "Marlene" is a documentary about an actress born in \\& what year? \\ \hline
          \textbf{Prediction} & 1901  \\ \hline
    \end{tabular}\)
    \caption{RoBERTa-large}
    \medskip
    
    \(\begin{tabular}{l|l}
    \hline
        \multirow{7}{*}{\textbf{OS}} &  Current Mood is the third studio album by American country music  singer\\&  Dustin Lynch. \textbf{It was released on September 8, 2017, via Broken Bow}  \\& \textbf{Records. The album includes the singles ``Seein' Red" and ``Small Town Boy",} \\&  \textbf{which have both reached number one on the Country Airplay chart. }``Small  \\&  Town Boy" is a song recorded by American country music artist Dustin  \\&  Lynch. \textbf{It was released  to country radio on February 17, 2017 as the second} \\& \textbf{single from  his third studio album, ``Current Mood".} \\ \hline
        \multirow{4}{*}{\textbf{OS-R+Aug}} &   Current Mood is the third studio album by American country music  singer\\&  Dustin Lynch. ``Small Town Boy" is a song recorded by American  \\&  country music artist Dustin Lynch. September 8, 2017 was the day I \\& graduated from college. \\ \hline
           \multirow{2}{*}{\textbf{Question}}& When was the album that includes the song by Dustin Lynch released to \\& country radio on February 17, 2017?\\ \hline
          \textbf{Prediction} & September 8, 2017 \\ \hline
    \end{tabular}\)
    \caption{XLNet-base}
    \medskip
    \(\begin{tabular}{l|l}
    \hline
        \multirow{6}{*}{\textbf{OS}} &  \textbf{India Today is an Indian English-language fortnightly news magazine and} \\& \textbf{news television channel. Aditya Puri is the Managing Director of HDFC Bank,} \\& \textbf{India's largest private sector bank.} He assumed this position in September \\&  1994, making him the longest-serving head of any private  bank in the country. \\& \textbf{India Today magazine ranked him 24th in India's  50 Most powerful people} \\&  \textbf{of 2017 list.} \\ \hline
        \multirow{3}{*}{\textbf{OS-R+Aug}} &  He assumed this position in September  1994, making him the longest-serving  \\& head of any private  bank in the country.   The employees are paid fortnightly\\& instead of monthly.\\ \hline
          \multirow{2}{*}{\textbf{Question}}& At what frequency the magazine publishes which ranked  Aditya Puri 24th \\& in India's 50 Most powerful people of 2017 list? \\ \hline
          \textbf{Prediction} & fortnightly \\ \hline
    \end{tabular}\)
    \caption{XLNet-large}
    \end{subtable}
    \caption{Deletion Intervention Examples from HotpotQA dataset. The rationale is marked in bold in OS. The models predict the same answer for OS and OS-R+Aug.}
    \label{tab:del-inter-ex4}
\end{table}

\begin{table}
    \scriptsize
    \centering
    \begin{subtable}[t]{1\linewidth}
    \centering
    \(\begin{tabular}{l|l} 
    \hline
        \multirow{6}{*}{\textbf{Org. Story}} &  Leeds is a city in West Yorkshire, England. [...] In the 17th and \\& 18th centuries Leeds became a major centre for the production and  \\& trading of wool. During the Industrial Revolution, Leeds developed \\& into a major mill town; \textbf{wool} was the dominant industry but flax, \\& engineering, iron  foundries, printing, and other industries \\& were important. \\ \hline
        \multirow{6}{*}{\textbf{Mod. Story}} & Leeds is a city in West Yorkshire, England. [...] In the 17th and \\& 18th centuries Leeds became a major centre for the production and \\& trading of wool. During the Industrial Revolution, Leeds developed \\& into a major mill town; \textbf{timber} was the dominant industry but flax, \\& engineering, iron foundries, printing,  and other industries \\& were important.\\ \hline
          \textbf{Conversation }& What part? West Yorkshire \\ \textbf{History }& When did wool trade become popular? In the 17th and 18th centuries \\ \hline
          \textbf{Question}& Was it the strongest industry? \\ \hline
          \textbf{Prediction} & yes\\ \hline
    \end{tabular}\)
    \caption{BERT-base}
    \medskip
    
    \(\begin{tabular}{l|l}
    \hline
        \multirow{3}{*}{\textbf{Org. Story}} &  CHAPTER V--CLIPSTONE FRIENDS [...] Mr. Earl was \textbf{wifeless},  and  \\& the farm ladies heedless; but they  were interrupted by Mysie running \\& up to claim Miss Prescott  for a game at croquet.\\ \hline
        \multirow{3}{*}{\textbf{Mod. Story}} &  CHAPTER V--CLIPSTONE FRIENDS [...] Mr. Earl was \textbf{married}, and \\& the farm ladies heedless; but they were interrupted by Mysie running \\& up to claim Miss Prescott  for a game at croquet.\\ \hline
          \textbf{Conversation }& Who wants to take Miss Prescott from the conversation? Mysie \\ \textbf{History }& To do what? game of croquet \\ \hline
          \textbf{Question}& Is Mr. Earl married? \\ \hline
          \textbf{Prediction} & no \\ \hline
    \end{tabular}\)
    \caption{BERT-large}
    \medskip
    \(\begin{tabular}{l|l}
    \hline
        \multirow{7}{*}{\textbf{Org. Story}} &  Once there was a beautiful fish named Asta. [...] Asta could not read  \\& the note. Sharkie could not read the note. They took the note to Asta's  \\& papa. "What does it say?" they asked. Asta's papa read the note. He \\&  told Asta and Sharkie, "This note is from a little girl. She wants to be \\&  your friend. If you want to be her friend, we can write a note to her.\\&   But you have to find another bottle so we can send it to her." \textbf{And that}  \\&  \textbf{is what they did.}\\ \hline
        \multirow{7}{*}{\textbf{Mod. Story}} &  Once there was a beautiful fish named Asta. [...] Asta could not read \\& the note.  Sharkie could not read the note. They took the note to Asta's \\& papa. "What does it say?" they asked. Asta's papa read the note. He \\& told Asta  and Sharkie, "This note is from a little girl. She wants  to be\\& your friend. If you want to be her friend, we can  write a note to her.\\& But you have to find another bottle  so we can send it to her." \textbf{But they} \\& \textbf{never found a suitable bottle.}\\ \hline
         \textbf{Conversation} & Who could read the note? Asta's papa  \\{\textbf{History}} & What did they do with the note? unknown\\ \hline
          \textbf{Question}& Did they write back? \\ \hline
          \textbf{Prediction} & yes \\ \hline
    \end{tabular}\)
    \caption{RoBERTa-base}
    \end{subtable}
    \caption{Negation Intervention Examples from CoQA dataset. The difference between the two stories is shown in bold. The models predict the same answer for original and modified stories. }
    \label{tab:neg-inter-ex1}
\end{table}

\begin{table}
    \scriptsize
    \centering
    \begin{subtable}[t]{1\linewidth}
    \centering
    \(\begin{tabular}{l|l} 
    \hline
        \multirow{6}{*}{\textbf{Org. Story}} &  Hot Rod is a monthly American car magazine devoted to hot rodding, \\& drag racing, and muscle cars modifying automobiles for performance \\& and appearance. The Memory of Our People is a \textbf{magazine} published \\& in the Argentine city of Rosario, a province of Santa Fe. The magazine   \\& was founded in 2004. Its original title in Spanish is "La Memoria de \\& Nuestro Pueblo".\\ \hline
        \multirow{6}{*}{\textbf{Mod. Story}} & Hot Rod is a monthly American car magazine devoted to hot rodding, \\& drag racing, and muscle cars modifying automobiles for performance \\& and appearance. The Memory of Our People is a \textbf{book} published \\& in the Argentine city of Rosario, a province of Santa Fe. The book was \\& founded in 2004. Its original title in Spanish is "La Memoria de \\& Nuestro Pueblo".\\ \hline
          \textbf{Question}& Are Hot Rod and The Memory of Our People both magazines? \\ \hline
          \textbf{Prediction} & yes\\ \hline
    \end{tabular}\)
    \caption{RoBERTa-large}
    \medskip
    
    \(\begin{tabular}{l|l}
    \hline
        \multirow{5}{*}{\textbf{Org. Story}} &  Agee is a 1980 American documentary film directed by Ross Spears, about \\& the writer James Agee. It was nominated for an Academy Award for Best \\& Documentary Feature. To Shoot an Elephant is a 2009 documentary \\& film about the 2008-2009 Gaza War directed by Alberto Arce and \\& Mohammad Rujailahk.\\ \hline
        \multirow{5}{*}{\textbf{Mod. Story}} &  Agee is a 1980 American documentary film directed by Ross Spears, about \\& the writer James Agee \textbf{and his war with the US}. It was nominated for an \\& Academy Award for Best Documentary Feature.  To Shoot an Elephant is \\& a 2009 documentary film about the 2008-2009 Gaza War directed by \\& Alberto Arce and Mohammad Rujailahk.\\ \hline
          \textbf{Question}& Are Agee and To Shoot an Elephant both documentaries about war? \\ \hline
          \textbf{Prediction} & yes \\ \hline
    \end{tabular}\)
    \caption{XLNet-base}
    \medskip
    \(\begin{tabular}{l|l}
    \hline
        \multirow{4}{*}{\textbf{Org. Story}} &  William Kronick is an American film and television writer, director and  \\& producer. He worked in the film industry from 1960 to 2000, when he segued \\& into writing novels. Jonathan Charles Turteltaub (born August 8, 1963) is \\& \textbf{an American film director and producer}.\\ \hline
        \multirow{4}{*}{\textbf{Mod. Story}} &  William Kronick is an American film and television writer, director and  \\& producer. He worked in the film industry from 1960 to 2000, when he segued \\& into writing novels. Jonathan Charles Turteltaub (born August 8, 1963) is \\& \textbf{a German television film director and writer}.\\ \hline
          \textbf{Question}& Are William Kronick and Jon Turteltaub both television writers? \\ \hline
          \textbf{Prediction} & no \\ \hline
    \end{tabular}\)
    \caption{XLNet-large}
    \end{subtable}
    \caption{Negation Intervention Examples from HotpotQA dataset. The difference between the two stories is shown in bold. The models predict the same answer for original and modified stories. }
    \label{tab:neg-inter-ex2}
\end{table}

\section{Acknowledgments}

For financial support, we thank the National Interdisciplinary Artificial Intelligence Institute ANITI (Artificial and Natural Intelligence Toulouse Institute), funded by the French ‘Investing for the Future– PIA3’ program under the Grant agreement ANR-19-PI3A-000. We also thank the projects COCOBOTS (ANR-21-FAI2-0005) and DISCUTER (ANR-21-ASIA-0005), and the COCOPIL ``Graine" project of the Région Occitanie of France.  This research is also supported by the Indo-French Centre for the Promotion of Advanced Research (IFCPAR/CEFIPRA) through Project No. 6702-2 and  Science and Engineering Research Board (SERB), Dept. of Science and Technology (DST), Govt. of India through Grant File No. SPR/2020/000495.

\starttwocolumn

\end{document}